\documentclass[10pt,journal,compsoc]{IEEEtran}
\usepackage{cite}
\usepackage{amsmath,amssymb,amsfonts}

\usepackage{changes}
\usepackage{amsmath}

\usepackage{algorithm}
\usepackage{algorithmic}
\usepackage{graphicx}
\usepackage{textcomp}
\usepackage{xcolor}
\usepackage{url}
\usepackage{subcaption}
\usepackage{arydshln}
\usepackage{amsthm}
\usepackage{multirow}
\usepackage{hyperref}
\usepackage{enumitem}

\newtheorem{definition}{Definition}
\newtheorem{lemma}{Lemma}
\newcommand\ct[1]{\textcolor{black}{#1}}
\newcommand\ic[1]{\textcolor{blue}{#1}}

\newcommand{\comments}[1]{}

\begin{document}
\title{NetFense: Adversarial Defenses against Privacy Attacks on Neural Networks for Graph Data}

\author{
        I-Chung~Hsieh,
        Cheng-Te Li,~\IEEEmembership{Member,~IEEE}
\IEEEcompsocitemizethanks{\IEEEcompsocthanksitem I-Chung Hsieh, Institute of Data Science, National Cheng Kung University, Tainan, Taiwan.
\IEEEcompsocthanksitem Cheng-Te Li, Institute of Data Science, National Cheng Kung University, Tainan, Taiwan.
}
\thanks{
}}

\markboth{IEEE Transactions on Knowledge and Data Engineering (TKDE)~2021}%
{Shell \MakeLowercase{\textit{et al.}}: Bare Demo of IEEEtran.cls for Computer Society Journals}

\IEEEtitleabstractindextext{%
\begin{abstract}
Recent advances in protecting node privacy on graph data and attacking graph neural networks (GNNs) gain much attention. 
The eye does not bring these two essential tasks together yet. 
Imagine an adversary can utilize the powerful GNNs to infer users' private labels in a social network. 
How can we adversarially defend against such privacy attacks while maintaining the utility of perturbed graphs? 
In this work, we propose a novel research task, adversarial defenses against GNN-based privacy attacks, and present a graph perturbation-based approach, NetFense, to achieve the goal.
NetFense can simultaneously keep graph data unnoticeability (i.e., having limited changes on the graph structure), maintain the prediction confidence of targeted label classification (i.e., preserving data utility), and reduce the prediction confidence of private label classification (i.e., protecting the privacy of nodes).
\comments{
We discuss how the perturbations of graph structure and node features influence GNNs, and devise a bi-optimization algorithm to generate the perturbed graphs that jointly maintain the performance of targeted label classification and lower down the prediction confidence of private label classification.}
Experiments conducted on single- and multiple-target perturbations using three real graph data exhibit that the perturbed graphs by NetFense can effectively maintain data utility (i.e., model unnoticeability) on targeted label classification and significantly decrease the prediction confidence of private label classification (i.e., privacy protection). Extensive studies also bring several insights, such as the flexibility of NetFense, preserving local neighborhoods in data unnoticeability, and better privacy protection for high-degree nodes.
\end{abstract}

}

\maketitle
\IEEEdisplaynontitleabstractindextext
\IEEEpeerreviewmaketitle
\IEEEraisesectionheading{\section{Introduction}\label{sec:introduction}}
\IEEEPARstart{G}{raph} data, such as citation networks, social networks, and knowledge networks, are attracting much attention in real applications. Graphs can depict not only node features but their relationships.
With the development of deep learning, 
graph neural networks (GNNs)~\cite{wu2019comprehensive} are currently the most popular paradigm to learn and represent nodes in a graph. 
GNN encodes the patterns from node features, aggregates the representations of neighbors based on their edge connections, and generates effective embeddings for downstream tasks, such as node classification, link prediction, and community detection. 
Typical GNN models include graph convolution-based semi-supervised learning~\cite{kipf2016semi}, and
generating relational features by incorporating input features with a columnar network~\cite{pham2017column}. In addition, graph attention is developed to estimate the contribution of incident edges~\cite{velivckovic2017graph}. The theory of information aggregation in GNNs has also discussed to enhance the representation ability~\cite{xu2018powerful}.
\begin{figure}[t]
  \centering
  \includegraphics[width=1.0\linewidth]{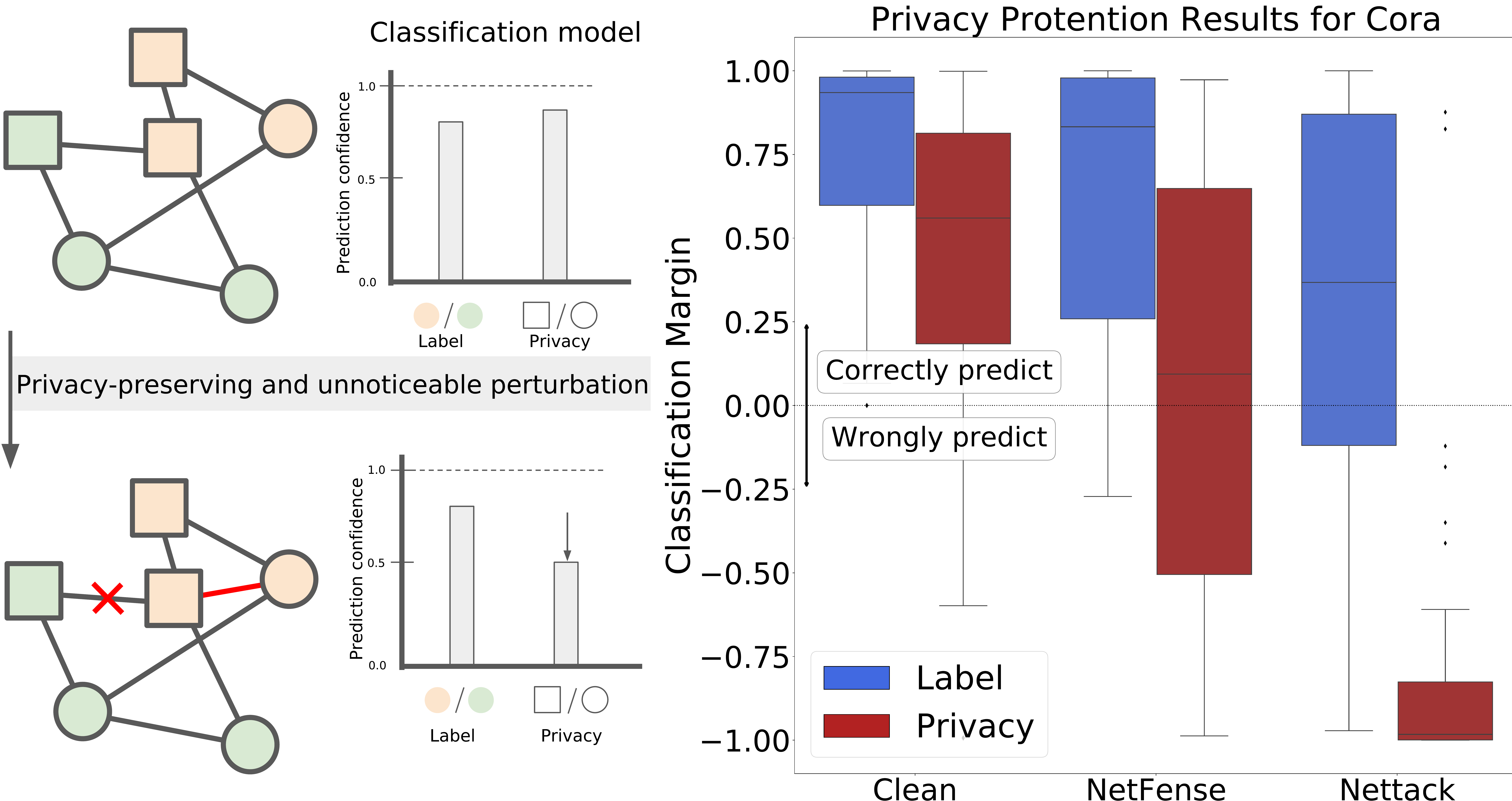}
  \caption{An elaboration of privacy-protected graph perturbation. \underline{Left}: we aim to perturb the given graph by removing an edge and adding a new one such that two requirements are satisfied: (1) The prediction confidence (y-axis) on private labels (i.e., square and circle) is lowered down, i.e., decreasing the risk of leaking privacy. (2) The prediction confidence on targeted labels (i.e., light green and yellow) is maintained, i.e., keeping the data utility. \underline{Right}: The proposed NetFense model can achieve such two requirements, compared to clean and perturbed data generated by Netteck~\cite{zugner2018adversarial}.}
  \label{fig:Intro}
  \vspace{-2.5em}
\end{figure}

Powerful GNNs make us concern about the disclosure of private information if the adversary considers private labels (e.g., gender and age) as the label. For example, 
while online social networks, such as Facebook, Twitter, and LinkedIn 
allow users to do privacy controls, partial data still have leakage crisis if users do not actively enable the privacy settings or agree with the access for external apps. As the adversary has partial data, GNNs can be trained to infer and acquire private information. For example, GNNs can be used to detect the visited location via user-generated texts on Twitter~\cite{rahimi2018semi}, and to predict age and gender using users' e-commerce records~\cite{chen2019semi}. 
Differential privacy (DP)~\cite{dwork2006calibrating} is a typical approach to add noise into an algorithm so that the risk of leaking private data can be lowered down. 
DPNE~\cite{xu2018dpne} and PPGD~\cite{ZhangDifferential}
devise shallow DP-based embedding models to decrease performances of link prediction and node classification.
However, since the original graph data cannot be influenced, such two models still lead to a high potential of risk exposure of private information.
We use Fig.~\ref{fig:Intro} (left) to elaborate the idea. Through a well-devised defense model, the graph is perturbed by removing one edge and adding another one. We expect that the new graph misleads the inference on private labels by decreasing the prediction confidence while keeping the data utility on target labels by maintaining the prediction confidence. 

\ct{In fact, an adversary can train a model based on the public-available data of some users' profiles in online social platforms, such as Twitter and Instagram. Not all of the users seriously care about their privacy. Hence, personal attributes and connections can be exposed due to two factors. First, some users may be not aware of privacy leaking when they choose to set some fields public. Second, some users do not care whether their private information is obtained by other people but are eager to promote themselves and maximize the visibility of themselves by proving full personal data. Data collected from such kinds of users allow the attackers to train the attack model. Therefore, we aim to find and fix the weaker parts from the data that can cause the risk of privacy exposure inferred by the attack model, and also to maintain the data utility. Then, the attackers would see the privacy-preserved data and cannot disclose the private labels of users by using the attack model.}

Nettack~\cite{zugner2018adversarial} is the most relevant study.
A gradient-based attack model is developed to perturb node features and graph structure so that the performance of a task (e.g., node classification) is significantly reduced. However, Nettack cannot work for privacy protection in two aspects. First, when the targeted private label is binary, the adversary can reverse the misclassified labels to obtain the true value if she knows there is some protection. Second, while Nettack can be used to defend against privacy attacks by decreasing the performance, it does not guarantee the utility of the perturbed data on inferring non-private labels. As shown in Fig.~\ref{fig:Intro} (right) conducted on real graph data, Nettack leads to misclassification on the private label, but fails to maintain the prediction confidence on the target label. Note that the y-axis is the classification margin, indicating the difference of prediction probabilities between the ground truth and the $2$-nd probable label. It can be also regarded as the prediction confidence of a model. Negative values mean higher potential to be identified as the $2$-nd probable label.

In this paper, we propose a novel problem of \textit{adversarial defense against privacy attack} on graph data. Given a graph, in which each node is associated with a feature vector, a targeted label (e.g., topic or category), and a private label (e.g., gender or age), our goal is to perturb the graph structure by adding and removing edges such that the privacy is protected and the data utility is maintained at the same time. To be specific, we aim at lowering down the prediction confidence on the private label to prevent privacy from being inferred by the adversary's GNNs, and simultaneously maintaining the prediction confidence on the targeted label to keep the data utility under GNNs when the data is released. This task can be treated as a kind of \textit{privacy defense}, i.e., defending the model attack that performs learning to infer private labels. 

We create Table~\ref{tab:att} to highlight the key differences between model attack (i.e., Nettack~\cite{zugner2018adversarial}) and our proposed privacy defense (i.e., NetFense) on graph data. First, since the model attack is performed by the adversary and the privacy defense is conducted by the data owner, their scope of accessible data is different. Second, as mentioned above, our problem is to tackle two tasks at the same time (i.e., fool the model on private labels and keep data utility), but model attack deals with only fooling the model on targeted labels. Third, in the context of privacy protection, decreasing the prediction accuracy on private labels cannot prevent them from being inferred if the private label is binary. The adversary can reverse the prediction results if she knows the existence of defense mechanism. Therefore, we reduce the prediction confidence as close as possible to $0.5$. Fourth, while model attack tends to make one or fewer nodes misclassified on targeted labels, privacy defense is expected to shield the private labels of more nodes from being accurately inferred. Last, both model attack and privacy defense need to ensure the perturbation on graph data is unnoticeable. Privacy defense further requires to achieve model unnoticeability, which is maintaining the performance of target label prediction using the perturbed graph under the same model (i.e., equivalent to maintain data utility). It is quite challenging to have a defense model that meets all of these requirements at the same time.

\begin{table}
\caption{Summary of differences between model attack and privacy defense on graph data in terms of who is doing the attack/defense (WHO), accessible data (\textbf{AD}), strategy (STG), perturbation objective (\textbf{PO}), non-noticeable perturbation (\textbf{NP}), number of tackled task (\textbf{\#Task}), and number of concerned targets (\textbf{\#Trg}). ``Pred-Acc'' and ``Pred-Confi'' are prediction accuracy and confidence. ``$\Join$'' is maintenance.}
\vspace{-0.5em}
\resizebox{\linewidth}{!}{%
\begin{tabular}{c|l|l}
\hline
                   & \textbf{Model Attack} (e.g., Nettack~\cite{zugner2018adversarial})             & \textbf{Privacy Defense} (i.e., NetFense)             \\ \hline
WHO     & the adversary                   & the data owner              \\\hdashline
AD & partial             & all                 \\\hdashline
STG     & fool model on target labels   & (1) fool model on private labels \\
        &                               & (2) keep utility on target labels \\\hdashline
PO      & Pred-Acc@target: $\downarrow$  & (1) Pred-Confi@privacy: $\leadsto 0.5$  \\
        &                                & (2) Pred-Confi@target: $\Join$ \\\hdashline
\#Task  & $1$ & $\geq2$  \\\hdashline
\#Trg  & fewer & more  \\\hdashline
NP          & data                         & data$+$model \\ \hline  
\end{tabular}
}
\label{tab:att}
\vspace{-1.5em}
\end{table}

To tackle the proposed privacy defense problem, we propose an adversarial method, NetFense\footnote{The code of NetFense can be accessed via the following Github link:
\url{https://github.com/ICHproject/NetFense/}}, based on the adversarial model attack. NetFense consists of three phases, including \textit{candidate selection}, \textit{influence with GNNs}, and \textit{combinatorial optimization}. The first phase ensures the perturbed graph is unnoticeable while the second and third phases ensure both privacy preservation and data utility (i.e., model unnoticeable) of the perturbed graph. 
We summarize the contribution of this paper as follows.
\begin{itemize}[leftmargin=*]
\item We propose a novel problem of adversarial defense against privacy attacks on graph data based on graph neural networks (GNNs). 
The goal is to generate a perturbed graph that can simultaneously keep graph data unnoticeability, maintain the prediction confidence of targeted label classification (i.e., model unnoticeability), and reduce the prediction confidence of private label classification (i.e., privacy protection).

\item We devise a novel adversarial perturbation-based defense framework, NetFense, to achieve the three-fold goal. We justify that perturbing graph structure can bring more influence on GNNs than perturbing node features, and prove edge perturbations can strike a balance between model unnoticeability and privacy protection.

\item We conduct experiments on single- and multi-target perturbations using three real datasets, and the results exhibit the promising performance of NetFense. Advanced empirical studies further give crucial insights, including our loss function is allowed to flexibly control the effect of model unnoticeability and privacy protection, the private labels of high-degree nodes can be better protected, the proposed PPR-based data unnoticeability strategy in NetFense can better preserve the local graph structure of each node, and perturbing graph structure is more effective to prevent private labels from being confidentially inferred than perturbing node features.
\end{itemize}

\textit{Paper Organization.} We review relevant studies in Sec.~\ref{sec-related}, and present problem statement in Sec.~\ref{sec-prob}. Sec.~\ref{sec-method} provides the technical details of NetFense, and Sec.~\ref{sec-exp} exhibits the experimental results. Sec.~\ref{sec-concl} concludes this work.
\vspace{-0.5em}


\section{Related Work}
\label{sec-related}
\vspace{-0.25em}
\textbf{Privacy-preserving Learning.} For non-graph data, Wang et al.~\cite{wang2018not} present a privacy-preserving deep learning model via differential privacy (DP). Beigi et al. ~\cite{BeigiRecommendation} devise a privacy-protected recommender system via bi-optimization. For graph data, existing studies develop shallow privacy-preserving models to achieve DP, such as DPNE~\cite{xu2018dpne} and PPGD~\cite{ZhangDifferential}. Besides, Zheleva et al. ~\cite{zheleva2009join}, Liu et al.~\cite{liu2016linkmirage} and Cai et al.~\cite{cai2016collective} protect the link privacy by obfuscating the social graph structure through neighborhood search and community detection. 
We compare privacy-preserving studies based on four aspects in Table~\ref{tab:PPprop} (top). (1) \underline{Data}: representing the input data type, including ``Continuous'' (e.g., images and user profiles) ``Graph.'' 
(2) \underline{Approach}: the way to ensure privacy protection in the generated data, including differential privacy (``DP''), establishing some privacy protection criteria for ``Optimization'', and utilizing data ``Statistics'' (e.g., degree or centrality) to change the graph structure so that the privacy is free from leaking. 
(3) \underline{Model}: a machine learning or deep learning model that is adopted by the adversary, including shallow models (``SM'') such as KNN, Bayesian inference and SVM, deep neural network (``DNN''), and graph neural network (``GNN''). 
(4) \underline{Goal}: the goals of the privacy-preserving studies are the same as ``Protect'' the private information. 
Although existing studies achieve some success, their goals are different from ours that relies on GNNs to simultaneously let the adversary misclassify private labels and maintain the utility of the perturbed graphs on targeted labels.

\begin{table}
\caption{Summary and comparison of related studies.}
\resizebox{\linewidth}{!}{%
\begin{tabular}{l|c|l|l|l|l}
\hline
                &   &Data &Approach  & Model&Goal         \\ \hline
\parbox[t]{2mm}{\multirow{4}{*}{\rotatebox[origin=c]{90}{Privacy}}}&~\cite{wang2018not} & Continuous & DP & DNN&Protect             \\ \cdashline{2-6}
&~\cite{BeigiRecommendation} & Continuous  & Optimization & DNN&Protect                \\\cdashline{2-6}
&~\cite{xu2018dpne, ZhangDifferential} & Graph & DP & SM&Protect                \\\cdashline{2-6}
&~\cite{zheleva2009join, liu2016linkmirage, cai2016collective} & Graph & Statistics & SM&Protect                 \\\hline
{/}&\textbf{NetFense} & {Graph} & {Statistics+Gradient}  & {GNN}&{Protect}\\\hline  
\parbox[t]{2mm}{\multirow{7}{*}{\rotatebox[origin=c]{90}{Adversarial}}}&~\cite{goodfellow2014explaining} & Continuous  & Gradient & DNN&Attack                \\\cdashline{2-6}
&~\cite{wang2019attacking} & Graph  & Optimization & DNN&Attack                \\\cdashline{2-6} 
&~\cite{dai2018adversarial, xu2019topology, zugner2019adversarial} & Graph  & Optimization & GNN&Attack                \\\cdashline{2-6}
&~\cite{zugner2018adversarial} & Graph  & Gradient & GNN&Attack                \\\cdashline{2-6} 
&~\cite{wu2019adversarial} & Graph  & Gradient & GNN&Attack                \\\cdashline{2-6} 
&~\cite{bhagoji2018enhancing, das2018shield} & Continuous  & Data Transformation & DNN&Defense                \\\cdashline{2-6} 
&~\cite{Entezari2020AllYN} & Graph  & Data Transformation & GNN&Defense                \\\cdashline{2-6} 
&~\cite{tang2020transferring} & Graph  & Optimization & GNN&Defense                \\\cdashline{2-6} 
&~\cite{zhu2019robust} & Graph  & Optimization & GNN&Defense                \\\cdashline{2-6} 
&~\cite{wu2019adversarial} & Graph  & Data Transformation & GNN&Defense                \\\hline 
\end{tabular}
}
\label{tab:PPprop}
\vspace{-2.0em}
\end{table}


\textbf{Model Attack on Graphs.} \ct{
The adversarial learning~\cite{chen2020survey} benefits the attacks on graphs from examining the robustness of the model via simulating the competitive model or data. On the one hand, the adversarial learning techniques can improve the classification performance like training with adversarial examples~\cite{pan2019learning}, which leads the model to learn the embeddings more precisely and robustly. On the other hand, the adversarial attack can fool the model to damage the prediction outcomes by discovering which parts in the graph structure are more sensitive to mislead the results.
} 
The typical model attack is FGSM~\cite{goodfellow2014explaining}, which is a gradient-based approach that utilizes small noise to create adversarial examples to fool the model. However, perturbing graph data is more challenging for the gradient-based method due to its discrete structure. 
Wang et al.~\cite{wang2019attacking} manipulate the graph structure by solving an optimization problem, but their method cannot deal with graphs with node features. 
Xu et al.~\cite{xu2019topology} also consider the optimization-based perturbation for attack and robustness problems, which only discuss un-specific targets. 
In addition, Dai et al.~\cite{dai2018adversarial} perturb graph data through Q-learning under GNN models. Nettack~\cite{zugner2018adversarial} is the first to perform adversarial attacks on attributed graphs and consider the unnoticebility of graph perturbations for GNNs. 
Then Z{\"u}gner et al.~\cite{zugner2019adversarial} incorporate meta learning with a bi-level optimization for GNN-based model attack on graphs. 
\ct{
On the other hand, the methods against model attack also gain much attention, including the techniques proposed by Bhagoji et al.~\cite{bhagoji2018enhancing} and Das et al.~\cite{das2018shield}, 
which focus on image data. 
For graph data, Entezari et al.~\cite{Entezari2020AllYN} discuss the low singular values of adjacency matrix can be affected by the attackers, and reconstruct the graph via low-rank approximation. 
Tang et al.~\cite{tang2020transferring} adopt transfer learning to incorporate the classifiers trained from perturbed graphs to build a more robust GNN against poisoning attacks. 
Zhu et al.~\cite{zhu2019robust} propose to leverage Gaussian distributions to absorb the effects of adversarial changes so that the GNN models can be trained more robustly.
Furthermore, Wu et al.~\cite{wu2019adversarial} revise the gradient-based attack to adaptively train robust GCN with binary node features, and develop a defense algorithm against the attack by making the GNN aggregation weights trainable to alleviate the influence of misleading edges. 
}

We again use Table~\ref{tab:PPprop} (bottom) to compare the studies of adversarial model attack on graphs. The meaning of some column names here are different from those of privacy-preserving learning.
\underline{Approach}: the adversarial method to fool or enhance data robustness, including ``Gradient''-based and ``Optimization''-based approaches to perturb graph data,
and ``data transformation'' to modify the graph structure (e.g., extracting the critical part via SVD or condensing the image to remove the noise).
\underline{Goal}: the adversary can ``Attack'' the model, and the defender can ``Defend'' against model attack. 
These studies perturb graph data for model attack, but have not connections with privacy protection on graphs. 
Our work is the first attempt to propose adversarial defense against privacy attack on graph data, in which the predictability of private labels is destroyed and the utility of perturbed graphs is maintained.

\section{Preliminaries}
\label{sec-prob}

\subsection{Definitions}
Let $G = (V, A, X)$ denote an attributed graph, where $V = \{ v_1, v_2, ..., v_N \}$ is the set of nodes, $A \in \{0, 1\}^{N\times N}$ is the adjacency matrix, and $X \in \{0, 1\}^{N\times d}$ is the feature matrix. The typical \textit{node classification} task assumes that nodes are annotated with known or unknown target labels, given by $C = \{c_1,c_2, ..., c_N\}\in \{0, 1\}^{N}$, and is to accurately predict the unknown labels of nodes by training a model (e.g., GNNs) based on $G$. 

The predictions of nodes' \textit{target labels} (e.g., topics and interests) and \textit{private labels} (e.g., age and gender) are treated as two node classification tasks: Target Label Classification (TLC) and Private Label Classification (PLC). Note that to simplify the problem, here we consider binary classifications for TLC and PLC. Different from the conventional adversary who aims at accurately performing PLC using $G$, in our scenario, the data owners (e.g., companies) are allowed to change and publish the data on their platforms by perturbing $G$ and generating $G'$, whose difference from $G$ is \textit{unnoticeable}, such that: (1) the adversary's PLC leads to worse performance, and (2) normal users' TLC performance based on $G'$ is as close as possible to that based on $G$. In this work, we play the role of the data owner (i.e., company).

\begin{definition}
Given graph $G = (V, A, X)$, the graph \textbf{perturbation} is the change of the graph structure $A$ or node features $X$ with budget $b$. Let the graph after perturbation be $G' = (V, A', X')$. The number of perturbations is $N_p = \sum|A-A'|/2+\sum|X-X'|$. We require $N_p \leq b$.
\label{def:perturbation}
\end{definition}

A perturbation is an action of changing a value in either $A$ or $X$ from $0$ to $1$ or from $1$ to $0$. For the perturbation on the adjacency matrix $A$, each change represents adding or removing an edge $(u,v)$ (e.g., updating one friendship). We assume that the change is symmetric since the graph is undirected.
For the perturbation on the feature matrix $X$, every change implies the adjustment of binary feature $x_i$ of a node (e.g., updating an attribute in the user profile).
Moreover, we limit the number of perturbations $N_p$ using a budget $b$ to have unnoticeable perturbations.

\begin{definition}
\textbf{Unnoticeable} perturbation indicates a slight difference between original graph $G$ and perturbed graph $G'$, and involves two aspects: \textbf{data} and \textbf{model}. The perturbation is \textbf{data-unnoticeable} if $|P_g(G)-P_g(G')|<\delta_g$ for a small scalar $\delta_g$ with respect to a given statistic $P_g$. Let $f_C: G \rightarrow \mathbb{R}^N$ is the trained GNN model that outputs the predicted TLC scores for all nodes. The perturbation is \textbf{model-unnoticeable} if $\sum_v |(f_C(G)-f_C(G'))|<\delta_c$ for a small scalar $\delta_c$.
\label{def:unnoticeable}
\end{definition}

The choice of graph statistic $P_g$ can be either degree distribution or covariance of features. GNN is used as the prediction model $f_C$ to measure the loss of information after perturbation. Note that model unnoticeability is equivalent to maintain the data utility for the normal usage of users. We formally elaborate the privacy-preserving graph that we perturb towards.

\begin{definition}
Let $P = \{p_1,p_2, ..., p_N\} \in \{0, 1\}^{N}$ be known/unknown private labels for all nodes in $G$. Also let $f_P: G \rightarrow \mathbb{R}^N$ be the trained GNN model that outputs the predicted PLC scores for all nodes. The perturbing graph $G' = (V, A', X')$ is \textbf{privacy-preserved} if $\sum_{v\in V} |\rho(f_P(G))-0.5|<\delta_p$ for a small scalar $\delta_p$, where $\rho: \mathbb{R} \rightarrow [0,1]$ is a scaling function.
\label{def:pp}
\end{definition}

The value of $\rho(f_P(G))$ is considered as the prediction confidence. Since the prediction is binary and the goal is to prevent private labels from being inferred, we want $\rho(f_P(G))$ to be as close to $0.5$ as possible. Such a requirement indicates that
the model $f_P(G)$ cannot distinguish which of the binary private labels is true. That is, it would become more difficult for the adversary to accurately obtain user's personal information.
Last, we combine all the above-defined to describe the proposed privacy-protected graph perturbation as follows. 

\begin{definition}
Given an attributed graph $G$ with target labels $C$ and private labels $P$, GNN models $f_C$ and $f_P$ can be trained for the prediction of $C$ and $P$, respectively. A perturbed graph $G'$ is a \textbf{privacy-preserved graph} from $G$ with budget $b$ if $G'$ is generated through $N_p$ \textbf{unnoticeable perturbations}.
\label{def:all}
\end{definition}

\ct{
Note that the actions of adding or removing edges need to rely on user will, and cannot be controlled by the company (i.e., the service provider). What the company can do is to provide ``suggestions'' for users, and meanwhile inform users the risk of privacy exposure by the attackers. There are two practical ways that the company can do to protect user privacy. First, when a user attempts to create a connection with another, the platform can display the privacy-leaking confidence for her reference, instead of enforcing her not to make friends. Some users who concern much about the privacy attacking can set a higher accessing level for their profile visibility. Second, the company can also provide users an option: automatically ``hide'' highly privacy-exposed connections from user profiles. Those who are afraid of being attacked can enable this option for privacy protection.
}

\subsection{Graph Neural Networks}
For the GNN models of $f_C$ and $f_P$ for semi-supervised node classification, we adopt Graph Convolutional Networks (GCN)~\cite{kipf2016semi}. The GCN information propagation to generate node embedding matrix $H$ from $l$ to $l+1$ layer is given by:  
$
    H^{(l+1)} = \sigma ( \tilde{D}^{(-\frac{1}{2})} \tilde{A} \tilde{D}^{(-\frac{1}{2})} H^{(l)} W^{(l)} ), 
$
where $\tilde{A} = A + I$ is the addition of adjacency and identity matrices.
$\tilde{D}$ is the recomputed degree matrix, $\tilde{D}_{ii} = \sum_{j} \tilde{A}_{ij}$. 
$H^{(l)}$ is the input representation of layer-$l$, and 
$H^{(0)} = X$ (using node features as input). $\sigma$ is the activation function and $W^{(l)}$ is the trainable weight. 

To have an effective prediction and avoid over-smoothing, we create GCNs with $2$ layers for node classification~\cite{wu2019comprehensive}. The output prediction probabilities $Z$ are:
\begin{equation}
    Z = f(G ; W^{(1)}, W^{(2)} ) = softmax( \hat{A} \sigma ( \hat{A} X W^{(1)} ) W^{(2)} ),
\label{eq:GCN2}
\end{equation}%
where $ \hat{A} =  \tilde{D}^{(-\frac{1}{2})} \tilde{A} \tilde{D}^{(-\frac{1}{2})}$ is the normalized adjacency matrix with self-loop.
The trainable weights $W^{(1)}, W^{(2)}$ are updated by cross-entropy loss $L$, given by:
$
    Loss = -\sum_{v \in V_{train}} \ln Z_{v, c_v},
$
where $c_v$ is the given label of $v$ in the training data $V_{train}$, and $Z_{v, c_v}$ is the probability of classifying node $v$ to label $c$. 

\section{The Proposed NetFense Framework}
\label{sec-method}
Our NetFense is an adversarial method that fits the transductive learning setting for GNNs. NetFense consists of three components, (a) candidate selection, (b) influence with GNNs models, and (c) combinatorial optimization.

\subsection{Candidate Edge Selection}
\label{sec-cand}
This component aims at ensuring data-unnoticeable perturbation.
Recall that we limit the number of perturbations $N_P$ using budget $b$ in Def.~\ref{def:perturbation}. 
We think each edge provides different contributions in achieving data-unnoticeable perturbation. 
A proper selection of edges to be perturbed can lead to better data-unnoticeability under the given budget.
Hence, we devise a perturbed candidate selection method, which selects candidate edges by examining how each edge contributes to data unnoticeability. Note that we perturb only edges, i.e., graph structure, because we have discussed in Sec.~\ref{sec-perturbcom} that the action of changing user attributes is more obvious and sensitive in real social networks. The number of perturbations accordingly becomes $N_P = \sum|A-A'|/2$.

We measure the degree of data unnoticeability using
Personalized PageRank (PPR) \cite{page1999pagerank} since it exhibits high correlation to the capability of GNN models~\cite{klicpera2018predict, bojchevski2019pagerank}.
Our idea is if the PPR scores of nodes before and after perturbations can be maintained, the perturbed graph is data-unnoticeable. In other words, PPR is treated as the graph statistic $P_g$ mentioned in Def.~\ref{def:unnoticeable}.
The PPR values on all nodes can be computed via ${{\pi}}_{r}^{(n)}={(1-\alpha){H}{\pi}}_{r}^{(n-1)} + \alpha {e}_r$, where ${{\pi}}_{r}^{(n)}$ is the probability vector from each starting node $r$ to all of the other nodes at $n$-th iteration, ${H} = {D}^{-1}{A}$ is the normalized adjacency matrix derived from ${A}$, ${D}$ is the degree matrix, $\alpha \in [0,1]$ is the restart probability, and ${e}_r$ is the one-hot encoding indicating the start node. 
The closed form of PPR matrix ${\Pi}$ is given by: ${\Pi} = \alpha (I-(1-\alpha) H)^{-1}$, where the entry ${\Pi}_{ij}$ denotes the stationary probability of arriving at node $j$ from node $i$. 

According to Def.~\ref{def:unnoticeable}, we set the $P_g={\Pi}$ and investigate the influence of edge perturbation $(u, v)$ on graph $G$. First, we simplify the symmetric effect of the undirected edge as one direction. We define the perturbation $(u, v)$ 
as an one-hot matrix $B$, where $B_{ij} =0$ if $(i, j) \neq (u, v)$, $B_{ij} =1- 2\cdot\mathbb{I}((u, v)\in G)$ if $(i, j) = (u, v)$ and the indicating function $\mathbb{I}((u, v)\in G) = 1$ if the edge $(i, j)$ is on graph $G$; otherwise, $\mathbb{I}((u, v)\in G) = 0$ if the edge $(i, j)$ is not on graph $G$. That is, $\mathbb{I}((u, v)\in G) = 1$ means an edge deletion $A'_{uv} = A_{uv}+(1- 2\cdot\mathbb{I}((u, v)\in G)) = A_{uv} -1$, and we would add a new edge to adjacency (i.e., $A'_{uv} = A_{uv} +1$) if $\mathbb{I}((u, v)\in G)= 0$.
Then, the graph after perturbation via $(u, v)$ can be denoted as a new adjacency matrix $A' = A+B$. We can derive the PPR score after perturbation via Lemma~\ref{lamma_inev} ~\cite{miller1981inverse}.
\begin{lemma}
\label{lamma_inev}
Let $M_1$ be an inevitable matrix, $M_2$ be a matrix with rank $1$ and $g = trace(M_2(M_1)^{-1})$. If $M_1 + M_2$ is inevitable, $g\neq -1$ and $(M_1+M_2)^{-1} = M_1^{-1} - (M_1^{-1}M_2 M_1^{-1})/(1+g)$.
\end{lemma}
To derive the new PPR score after the perturbation, we let $M_1 = I-(1-\alpha)H$ and $M_2 = -(1-\alpha)H'$, where $H' = D^{-1}B$, from which we suppose the degree shift is small, i.e., $D'\approx D$, for the degree matrix $D'$ of $A'$. The change of one entry of $A$ would affect only $D_{uu}$, i.e., $D_{uu} \pm 1$. Hence, we can 
minimize the difference between $D$ and $D'$ based on the given budget. Note that the normalized matrix $H$ with non-negative and bounded entries satisfies the condition of Neumann series and is invertible. Besides, we prevent the graph from being disconnected by not considering nodes with degree $1$ for perturbation, to ensure $H'$ is invertible. In detail, by employing PPR as the statistic $P_g$, we derive the formula that depicts the difference of PPR score $P_{g}$ for from any nodes $i$ to $j$ by perturbing the directed edge $u \to v$, denoted by $\Delta_{u \to v} P_{g}[i,j]$, as follows:
\begin{equation}
\begin{aligned}
     \Delta_{u \to v} P_{g}&[i,j] = \{P_g(G') - P_g(G)\}_{ij} \\
     &= \{ \alpha (M_1+M_2)^{-1} - \alpha M_1^{-1} \}_{ij} \\
     &= -\alpha\{(M_1^{-1}M_2M_1^{-1})/(1+g)\}_{ij} \\
     &= -\alpha\{(M_1^{-1} (- (1-\alpha)D^{-1}B ) M_1^{-1})/(1+g)\}_{ij} \\
     &= \alpha d_u^{-1}(1-\alpha) \{M_1^{-1} B M_1^{-1}\}_{ij}/(1+g)) \\
     &= \alpha c' \{(M_1^{-1})_{*u} (M_1^{-1})_{v*}\}_{ij}/(1+g) \\
     &= \alpha c' (M_1^{-1})_{iu} (M_1^{-1})_{vj}/(1-c'(M_1^{-1})_{vu}) \label{eq:PPR}
\end{aligned}%
\end{equation}
where $c' =  b_s d_u^{-1}(1-\alpha)$ and $b_s = sign(B_{uv})$. We can directly derive the quadratic term and trace of $M_1$ by selecting related column $(M_1^{-1})_{*u}$ and row $(M_1^{-1})_{v*}$ since $B$ is a one-hot matrix and only values with the corresponding indexes are took account. To reduce the destruction of graph structure, we consider the perturbation of edge $u \to v$ with the lower absolute value of $\Delta_{{u \to v}} P_{g}[i,j]$ as the candidate. In other words, $\Delta_{{u \to v}} P_{g}[i,j]$ is regarded as the influence on PPR values by perturbing edge $u \to v$.
Regarding $(M_1^{-1})_{iu}$ and $(M_1^{-1})_{vj}$ in the numerator, the strength of such interaction effect depends on the degree of the original relationship among neighbors of $u$ and $v$. Higher values of $(M_1^{-1})_{iu}$ and $(M_1^{-1})_{vj}$ reflect the larger influence for perturbation $(u,v)$. Regarding $(M_1^{-1})_{vu}$ in the denominator, such an individual effect comes from the reverse direction $(v,u)$, and is relevant to the action of adding or deleting. 

Since the PPR derived by iterating the passing probability through nodes' directed edges, the influence of the perturbation results from not only the paths $u' \leadsto u$ and $v \leadsto v'$ for any nodes $u', v' \in V$, but the effect of its opposite direction ($v \to u$). For the case of edge addition, we have $sign(-c'(M_1^{-1})_{uv}) = sign(-b_s) = sign(-(1-0))<0$, which means lower $(M_1^{-1})_{vu}$ has lower $\Delta_{{u \to v}} P_{g}[i,j]$. That is, for an directed edge $v \to u$ that some paths can pass through, the addition of edge $u \to v$ would make it a bi-directed one, and bring the increasing flow for PPR values. On the other hand, for edge deletion, we have $sign(-c'(M_1^{-1})_{uv}) = sign(-b_s) = sign(-(1-2))>0$, indicating that larger $(M_1^{-1})_{vu}$ decreases the change of PPR. That said, if edge $v \to u$ can support the unnoticeable structural change in terms of PPR, the deletion of edge $u \to v$ would not cause much destruction of PPR flow from nodes $v$ to $u$. However, we need not to consider edge direction. We assume the graph is undirected, and the edge change is symmetric. Hence, we need to revise the denominator term. 
We combine all influence of PPR from any nodes $i$ to $j$ by perturbing $(u,v)$, denoted by $\Delta_{u \leftrightarrow  v}P_{g}$, in which the other edge direction $v \to u$ is took into account, given by:
\begin{equation}
\begin{aligned}
     \Delta_{u \to v}P_{g} &= \sum_{(i,j)} \Delta_{u \to v} P_{g}[i,j] \label{eq:oneside}\\
     &= \sum_{(i,j)} \alpha c' (M_1^{-1})_{iu} (M_1^{-1})_{vj}/(1+c'(M_1^{-1})_{uv}) \\
     &= \sum_{i} c' (M_1^{-1})_{iu} /(1+c'(M_1^{-1})_{uv})\\
\end{aligned}%
\end{equation}
\begin{equation}
\begin{aligned}
     \Delta_{u \leftrightarrow  v}P_{g} &= |\Delta_{u \to v}P_{g} + \Delta_{v \to u}P_{g}|,\label{eq:twoside}
\end{aligned}%
\end{equation}
where $\sum_{j}\alpha  (M_1^{-1})_{vj} = 1$ since the row of the PPR marix $\Pi_v$ represents all probabilities for node $v$ to each node. 
Considering the adjustment of edge direction, we can simply focus on the PPR score regarding $-(M_1^{-1})_{uv}$ 
to achieve the data unnoticeability. The new formula ensures the low influence of perturbation $\Delta_{u \to v}P_{g}$ by deleting edges with lower PPR (i.e., $sign(c'(M_1^{-1})_{uv})<0$) and adding edges with higher PPR (i.e., $sign(c'(M_1^{-1})_{uv})>0$). In other words, we add the edges with higher $(M_1^{-1})_{uv}$, and delete the edges with lower $(M_1^{-1})_{uv}$. Both the PPR influence on all node pairs and the symmetric effect are considered as the measurement of candidate selection.

The formula above is to discuss the data unnoticeablity of a single edge perturbation (i.e., $N_p=1$). In practice, we require multiple edge perturbations. Here we aim at selecting multiple edge candidates being perturbed to approximate our original setting of data unnoticeability in $|P_g(G)-P_g(G')|<\delta_g$ with $N_p>1$ in Def. \ref{def:unnoticeable}. We find every candidate node pair by setting an upper bound $\tau$, and consider node pair $e=(u, v)$ as our candidate if it satisfies the data unnoticeability, i.e., $|P_g(G)- P_g(G'))| = \Delta_{u \leftrightarrow  v}P_{g}<\tau$, where $G' = G \pm e$, and the threshold $\tau$ is a given hyperparamerter. We collect all selected node pairs into the candidate set $Cand = \{(v,u) | (\Delta_{u \leftrightarrow  v}P_{g}<\tau) \wedge (v,u \in V) \}$ for the following two components.

\subsection{Influence with GNNs}
\label{sec-infgnn}
To achieve both goals of model unnoticeablility (i.e., data utility) in Def. \ref{def:unnoticeable} and privacy preservation in Def. \ref{def:pp}, we need to investigate how the perturbation affects the trained GCN model. In this section, we first discuss the difference between perturbing graph structure and perturbing node features. Then we elaborate how to generate the perturbed graph structure using edge additions and removals.

\subsubsection{Perturbation Comparison between Graph Structure and Node Features on GNNs}
\label{sec-perturbcom}
Before devising an approach to select better candidates of perturbations, we need to understand to what extent perturbing graph structure and node features contribute to the prediction results of GNN. We will verify that perturbing structure has more significant influence than perturbing node features in terms of GNN prediction probabilities.

\textbf{GCN Simplication.} We simplify the activation function in GNNs to better discuss the prediction difference between graph structure and node features. Given the adjacency matrix $A$ and the feature matrix $X$, the output prediction probabilities $Z'$ of a $2$-layers GCN without activation functions is given by:
\begin{equation}
    Z'_{i}(A,X)= \left(\hat{A}^2 X W^{(1)}  W^{(2)}\right)_i = \sum_{v_j \in N_{G}^{(2)}(v_i)} \hat{A}_{ij}^2 x_j^\top W' ,
\label{eq:GCNs}
\end{equation}%
where $W' = W^{(1)}W^{(2)}$ learned during pre-training. $N_{G}^{(2)}$ is the set containing the first- and second-hop neighbors derived from the $2$-nd order transition matrix for graph $G$, in which $\hat{A}_{ij}^2 = 0$ if the minimum distance between nodes $i$ and $j$ is greater than $2$.
Omitting the activation functions alleviates the computation for the propagation of node features with learned weights, without losing much information. We find the simplified output still involves the aggregation of propagated features from a node's first- and second-hop neighbors using trained weights. In other words, the simplified GCN aggregates the first and second order neighbors at the same layer, and the weights learned from GCN's second layer produce the deviation since we ignore the regulating by activation.

To evaluate the influence of perturbation, we compute $Z_i'(A',X')=(\hat{A'}^2 X' W^{(1)} W^{(2)})_i$ with perturbed graph $A'$ and feature $X'$, and compare it with $Z'_{i}(A,X)$. We analyze how prediction probabilities are affected by perturbing node features and graph structure, respectively. Here we assume every perturbation is performed on either one feature or one edge at one time. 

\textbf{Feature Perturbation.} Considering a specific entry $x_{kl}$ depicting the feature $l$ of node $k$, we denote the perturbation by: $\Delta(x_{kl}): x_{kl} \to (1-x_{kl})$, which indicates such a binary feature becomes opposite. We derive the difference of $Z'$ under $\Delta( x_{kl})$, denoted as $\epsilon_{\Delta(x_{kl})}Z'$, given by:
\begin{align*}
     \epsilon_{\Delta(x_{kl})}Z' 
     &= Z'_i(A,X') - Z'_i(A,X)\\ 
     &= \left(\hat{A}^2 X' W'\right)_i - \left(\hat{A}^2 X W'\right)_i\\
     &= \left(\hat{A}^2 (X'-X) W'\right)_i \\
     &= \sum_{v_j \in N_{G}^{(2)}(v_i)} \hat{A}_{ij}^2 \left({x'}_j-x_j\right)^\top W' \\
     &= \hat{A}_{ik}^2 \left({x'}_k-x_k\right)^\top W'\\
     &= \hat{A}_{ik}^2 h_{kl}^\top W'.
\label{eq:GCN2sx}
\end{align*}%
where $h_{kl}\in \mathbb{R}^d$ is the one-hot encoding vector with value $1-2x_{kl}$ for feature $l$ 
, and $h_{kg}=x'_{kg}-x_{kg}=0$ for $g \neq l$.
The result $\epsilon_{\Delta(x_{kl})}Z'=|\hat{A}_{ik}^2h_{kl}^\top W'|$ implies that the feature perturbation only influences the prediction outputs for $\hat{A}_{ik} \neq 0$, i.e., nodes nearby the target $k$.

\textbf{Structure Perturbation.} A structure perturbation is to add or remove an edge $(k,m)$ between nodes $k$ and $m$, denoted as $\Delta(e(k,m)): A_{km} \to (1-A_{km})$. Structure perturbation affects the adjacency matrix $A$ in deriving $Z'$. We derive the difference of $Z'$ under $\Delta(e(k,m))$, denoted as $\epsilon_{\Delta(e(k,m))}Z'$, as follows:
\begin{align*}
     \epsilon_{\Delta(e(k,m))}&Z' =
     Z'_i(A',X) - Z'_i(A,X) \\ 
     &= \left(\hat{A'}^2 X W'\right)_i - \left(\hat{A}^2 X W'\right)_i \\
     &= \left(\left(\hat{A'}^2 - \hat{A}^2\right) X W'\right)_i \\
     &= \left(\sum_{v_j \in N_{G'}^{(2)}(v_i)} \hat{A'}_{ij}^2 x_j^\top - \sum_{v_j \in N_{G}^{(2)}(v_i)} \hat{A}_{ij}^2 x_j^\top\right)  W',
\end{align*}%
where $\hat{A'}_{ij}^2$ is the normalized adjacency matrix with self-loop under ${\Delta(e(k,m))}$, and $N_{G'}^{(2)}(v_i)$ is the new neighborhood of node $v_i$ in the perturbed graph $G'$. The structure perturbation influences a node's neighbors for the aggregation and weights due to the changed $N_{G'}^{(2)}(v_i)$ and $\hat{A'}^2$. Such change results from increasing or decreasing the minimum distance between nodes $k$ and $m$, which further affects the numbers of their first- and second-hop neighbors.

To sum up, by looking into a particular node $k$ that involves in the perturbation of feature ${\Delta(x_{kl})}$ or edge ${\Delta(e(k,m))}$, we can understand which parts in GNN prediction $Z'$ are influenced. For feature perturbation, ${\Delta(x_{kl})}$ brings $|\hat{A}_{ik}^2h_{kl}^\top W'|$ for $\hat{A}_{ik}^2 \neq 0$. Such influence locates at node $k$'s specific feature and its first two hops' neighbors. On the other hand, for structure perturbation, ${\Delta(e(k,m))}$ produces $|(\sum_{v_j \in N_{G'}^{(2)}(v_i)} \hat{A'}_{ij}^2 x_j^\top - \sum_{v_j \in N_{G}^{(2)}(v_i)} \hat{A}_{ij}^2 x_j^\top)|$. Its influence range can cover the aggregation terms ($N_{G'}$), the adjacency weights ($\hat{A'}$) for nodes $k$ and $m$, and their first two hops' neighbors. Consequently, we can apparently find that structure perturbation affects wider on the GNN model. This suggests that perturbing edges takes more effect.

\subsubsection{Calcuating the Updated Graph Structure}
Based on the above analysis, to estimate how the perturbation of edge $(k,m)$ affects the predicted probability on a target node $k$, we need to recalculate $\hat{A}$. Such a recalculation has a high computational cost.
Since the $2$-layer GNN-based prediction relies on only the first- and second-hop neighbors of the target node $k$, i.e., most elements in $\hat{A}$ are zero, we can have an incremental update.  
In perturbing edge $(k,m)$, we extend Theorem 5.1 in Netteck~\cite{zugner2018adversarial} to consider the row $k$ of $\hat{A}^2$ as well as the computation of $\hat{A'}_{ij}^2$ in term of $\hat{A}_{ij}^2$ to have an efficient update. The updated $\hat{A'}_{ij}^2$ is given by:
\begin{equation}
\label{adj_update}
\begin{split}
     \hat{A'}_{ij}^2 & = \frac{1}{\sqrt{\tilde{d'}_i \tilde{d'}_j}}
     \Biggl[
     \sqrt{\tilde{d}_i \tilde{d}_j}\hat{A}_{ij}^2 
     +\left(
     \frac{\tilde{a'}_{ij}}{\tilde{d'}_i}
     -\frac{\tilde{a}_{ij}}{\tilde{d}_i}
     +\frac{\tilde{a'}_{ij}}{\tilde{d'}_j}
     -\frac{\tilde{a}_{ij}}{\tilde{d}_j}
     \right)\\
    &+\left(
    \frac{{a'}_{ik}{a'}_{kj}}{\tilde{d'}_k}
    -\frac{{a}_{ik}{a}_{kj}}{\tilde{d}_k}
    +\frac{{a'}_{im}{a'}_{mj}}{\tilde{d'}_m}
    -\frac{{a}_{im}{a}_{mj}}{\tilde{d}_m}
     \right)
     \Biggr],
\end{split}%
\end{equation}
where the lowercase notations are the elements of their corresponding matrix forms: the self-loop adjacency matrix $\tilde{A}_{ij} := \tilde{a}_{ij}$, the original adjacency matrix $A_{ij}:=a_{ij}$, the self-loop degree matrix $\tilde{D}_{ii}:=\tilde{d}_{i}$, and the updated matrix $A'_{ij} := a'_{ij}$ being positioned within ``$\left[\; \right]$.''
Updated elements $\tilde{a'}$, $a'$, $\tilde{d'}$ can be obtained via: 
\begin{align*}
    \tilde{d'_i} & = \tilde{d_i} + \mathbb{I}(i \in \{k, m\})(1-2\cdot a_{km})\\
    {a'_{ij}} & = {a_{ij}} + \mathbb{I}((i, j) = (k, m) )(1-2\cdot {a_{ij}})\\
    \tilde{a}'_{ij} & = \tilde{a}_{ij} + \mathbb{I}((i, j)  = (k, m))(1-2\cdot {a_{ij}}),
\end{align*}%
where the indicating function $\mathbb{I}(i \in \{k, m\}) =1$ is used to update node $k$'s degree if $i=k$ or $i=m$ (e.g., $\tilde{d'_k} = \tilde{d_k} + (1-2\cdot a_{km})$ for $i=k$); else $\mathbb{I}(i \in \{k, m\}) = 0$ implies no degree change. On the other hand, $\mathbb{I}((i, j) = (k, m)) = 1$ indicates the changes between ${a}_{ij}$ and ${a}'_{ij}$ and between $\tilde{a}_{ij}$ and $\tilde{a}'_{ij}$ are only applied when $i=k$ and $i= m$ (e.g., ${a'}_{km} = {a}_{km} + (1-2\cdot a_{km})$ for the case ${a'}_{km}$); else $\mathbb{I}((i, j) = (k, m)) = 0$. 

Equation~\ref{adj_update} consists of three parts, including the original adjacency matrix $\hat{A}_{ij}^2$, the direct adjustment in the first bracket, and the indirect adjustment in the second bracket, in which $j \in N_{G}^{(2)}$ that reflects the influence of perturbation covers the first two neighbors of node $k$. The first adjustment comes from the direct influence of perturbation ${\Delta(e(k,m))}$ and node $k$'s first-order neighbors. The second-order neighbors of node $k$ suffer from the indirect adjustment via its first-order neighbor $m$. Equation~\ref{adj_update} exhibits the influence of perturbation on the first- and second-order neighbors, and provides guidance for us to select more effective perturbation candidates in the next phase. 

\subsection{Perturbations via Combinatorial Optimization}
\label{sec-opt}
We generate the best combination of edges being perturbed, i.e., the perturbed graph, by a combinatorial optimization. We iteratively find the set of edge candidates $Cand$ that satisfies the data unnoticeability described in Sec.~\ref{sec-cand}, then select an edge (from $Cand$) at one time that leads to the best model unoticeability (i.e., data utility) and privacy preservation discussed in Sec.~\ref{sec-infgnn}. The selection of edges being perturbed will stop when either running out of the budget $b$ or no candidates in $Cand$. Below we first present our NetFense algorithm that generates the perturbed edges when the adversary aims to attack the privacy of a single node $v$. Then we elaborate the extension from single-target attacks to multi-target attacks.

For single-target attacks, we consider the protection of the private label for one node $v\in V$. We can derive the corresponding node-pair candidates $Cand(v) = \{(v,u) | (\Delta_{u \leftrightarrow v}P_{g}<\tau) \wedge (u \in V) \}$ at each time of perturbation selection. That is, the perturbation candidates are node pairs whose one end is the target node $v$. Some candidate node pairs are selected to create new edges while some are removed from the graph. 
The objective for perturbation selection consists of two parts. To ensure model unnoticeability, we devise the objective $O_C$ for target label classification, given by $O_C = [|f_C(G)-f_C(G')|]_v$, where $f_C = \hat{A}^2 X W_C$. The objective $O_C$ aims to minimize the classification error of node $v$. To protect node privacy, we design the objective $O_P$ for private label classification, i.e., minimizing $O_P = |[\rho(f_P(G))]_v-0.5|$, where $f_P = \hat{A}^2 X W_P$. Both $f_C$ and $f_P$ are simplified pre-trained GCN models, and $W_C$ and $W_P$ are learnable weights.
We formulate the final loss function to solve the bi-tasks, i.e., minimizing both $O_C$ and $O_P$, as follows:
\begin{equation}
    \mathcal{L}(G', W_C, W_P , v) = \frac{|[\hat{A'}^2XW_P]_{v}^{p_1} - [\hat{A'}^2XW_P]_{v}^{p_2} |^{a_d}}{ ([\hat{A'}^2 XW_C]_{v}^{\check{c}})^{a_m} },
\label{eq:Score}
\end{equation}%
where $p_1$ and $p_2$ are two classes of the specified binary private label, $[\hat{A'}^2XW_P]_{v}^{p_1}$ is the predicted score of node $v$ on private class $p_1$, $\check{c}$ is the predicted target label by the pre-trained GCN, and $[\hat{A'}^2 XW_C]_{v}^{\check{c}}$ is the predicted score of node $v$ on a particular predicted target class $\check{c}$. In addition, ${a_d}$ and ${a_m}$ are balancing hyperparameters that control the strength of maintaining the performance of target label classification (for model unnoticeability) and reducing the performance of private label classification (for privacy protection). Below we elaborate the loss function. Larger ${a_d}$ and smaller ${a_m}$ implies concentrating more on privacy protection than model unnoticeability. We will analyze how ${a_d}$ and ${a_m}$ affect the performance in the experiments.

For privacy protection, as depicted by the numerator in Equation~\ref{eq:Score}, since the private label is binary (i.e., only two classes $p_1$ and $p_2$), the score of one class approaches $0.5$ implies scores of both classes are close. Hence, we can rewrite the objective $O_P$ as: $|[\hat{A'}^2XW_P]_v^{p_1} - [\hat{A'}^2XW_P]_v^{p_2}|$. The scaling function $\rho$ is skipped here because minimizing the difference with scaling can be approximated by that without scaling. Lower values of rewritten $O_P$ indicate that the perturbed graph leads to lower confidence in disclosing private labels of the target node since two classes $p_1$ and $p_2$ are indistinguishable.

To maintain model unnoticeability, as depicted by the denominator in Equation~\ref{eq:Score}, since we cannot know the ground-truth target labels for all nodes, what we can utilize is the label $\check{c}$ predicted by the pre-trained GCN model.  
Since the predicted score $f_C(G)$ on the target label using the original graph is a constant for the fixed input and parameters, and to strongly prevent edges that contributed to target label classification from being selected as perturbation candidates, we rewrite the objective $O_C$ from minimizing $|[f_C(G)-f_C(G')]_{v}^{\check{c}}|$ to approximately maximize $[f_C(G)]_{v}^{\check{c}}$, which can be further transformed to minimize $([f_C(G)]_{v}^{\check{c}})^{-1}$. Higher values of $[f_C(G)]_{v}^{\check{c}}$ tend to maintain the performance of target label classification.

We summarize the proposed single-target adversarial defense framework, NetFense, in Algorithm~\ref{alg:algorithm}. We first use the original graph $G=(V,A,E)$ to pre-train two GCN models, $f_C$ and $f_P$, so that the loss in Equation~\ref{eq:Score} can be constructed using $W_C$ and $W_P$. Then by assuming every node pair $(u,v)$ is perturbed, we compute all influence of PPR between nodes, $\Delta_{u \leftrightarrow  v}P_{g}(G)$, which is used to limit the number of possible candidates for perturbation of the given target. 
In selecting each structural perturbation until either running out of the budget $b$ or no appropriate candidates (line 4-15), we rely on calculating the loss in Equation~\ref{eq:Score} to choose the best candidate at one time. We generate the perturbation candidates $Cand(v)$ based on threshold $\tau$ (line 5), accordingly produce the perturbed graphs by adding and removing every candidate in $Cand(v)$ (line 7-8), and find the perturbed one $G'^{*(t)}$ that minimizes Equation~\ref{eq:Score} (line 9).

The NetFense algorithm can be extended for multi-target adversarial defense by repeating single-target NetFense (Algorithm~\ref{alg:algorithm}) on different targets. That said, we can randomly choose a node from the target set without replacement, and then apply NetFense algorithm in an iterative manner. The perturbed graph for the current target will be utilized to generate the new perturbed graph of the next target.

\begin{algorithm}[tb]
\caption{NetFense}
\label{alg:algorithm}
\textbf{Input}: Graph $G = (V, A, X)$, single target $v$, threshold $\tau$, perturbation budget $b$, and hyperparameters $({a_d},{a_m})$\\
\textbf{Output}: Perturbed Graph $G' = (V, A', X)$\\
    \begin{algorithmic}[1] 
    \STATE Train two GNNs $f_C$ and $f_P$ using $G$ to get $W_C$ and $W_P$\\
    \STATE Compute $\Delta_{u \leftrightarrow  v}P_{g}(G)$ for all possible node pairs\\
    \STATE Let $G^{(0)} = G$ and $t=0$
    \WHILE{$\sum (A^{(0)}-A^{(t)}) \leq b$}
    \STATE  $Cand(v) = \{(v,u)  | (\Delta_{u \leftrightarrow  v}P_{g}<\tau) \wedge (u \in V) \}$ 
    \IF {$|Cand(v)| > 0$}
    \STATE $\{G'^{(t)}\}=\{G^{(t)}\pm(k, l)| (k, l) \in Cand(v)\}$
    \STATE Update $\hat{A'}^2$ for every graph in $\{G'^{(t)}\}$ by Eq.~\ref{adj_update}
    \STATE $G'^{*(t)} = \arg\min_{G'\in \{G'^{(t)}\}}\{\mathcal{L}(G', W_C, W_P , v)\}$
    \ELSE
    \STATE break
    \ENDIF
    \STATE $G^{(t+1)} = G'^{*(t)}$
    \STATE $t = t+1$
    \ENDWHILE
    \STATE  \textbf{return} $G^{(t)}$
    \end{algorithmic}
\end{algorithm}

\section{Experiments}
\label{sec-exp}
\subsection{Evaluation Settings}
Three real-world network datasets, including Cora, Citeseer~\cite{zugner2018adversarial}, and PIT~\cite{zhao2006entity}, are utilized for the evaluation. The statistics is shown in Table \ref{tab:data}. Cora and Citeseer are citation networks with 0/1-valued features, corresponding to the absence of words from the dictionary, and their TLC labels are paper categories. PIT depicts the relations between terrorists, along with anonymized features. The label ``colleague'' is used as the TLC task. 
\ct{Since the private label is not defined in these datasets, we regard one feature as the privacy label to simulate that few users do not set their sensitive attribute (e.g., hometown location) invisible. That is, we assume the attack aims to encroach on the target feature (e.g., hometown location) as the PLC label by building the model based on social relationships, other features, and a few visible PLC labels.}
To let the adversary have stronger attack capability, in each dataset, we select the binary attribute with the most balanced 0/1 ratio and having at least $0.6$ accuracy to be the PLC label. Besides, we will also study the correlation between TLC and PLC labels. By adopting \textit{Cohen's kappa coefficient} $\kappa$~\footnote{\url{https://en.wikipedia.org/wiki/Cohen\%27s_kappa}}, we can measure the correlation between targeted and private labels. Higher $\kappa$ (close to 1) indicates higher agreement between PLC and TLC labels, and means the performance of TLC has higher potential to be influenced when we perform perturbations for PLC. Here, we report the $\kappa$ coefficient values between TLC and PLC labels for the choices of our private labels: $-0.13$, $0.01$ and $-0.14$, for Citeseer, Cora, and PIT, respectively. We further discuss label correlation between PLC and TLC in Sec.~\ref{sec-exp-adv}.
For the data processing, we split each dataset into training, validation, and testing sets with ratio $0.1$, $0.1$, and $0.8$ for semi-supervised learning.

In the hyperparameter settings, we train the 2-layer GCN with $16$-dim hidden layer tuned via the validation set. We set the PPR restarting probability $\alpha = 0.1$, balancing values in loss function Eq.~\ref{eq:Score}: $a_d = 2$ and $a_m = 1$. For candidate selection, we set the quantile $\tau=0.9$, indicating that we retain node pair $(u^*, v^*)$ if its $\Delta_{u^* \leftrightarrow  v^*}P_{g}(G) \leq 0.9$ quantile of $\{\Delta_{u \leftrightarrow v}P_{g}(G) | u,v\in V \}$. That said, we exclude extremely influential node pairs. 

\begin{table}
\centering
\caption{Summary of data statistics.}
\begin{tabular}{lcccc}
\hline
Dataset  & \#nodes & \#edges & density &  \#features\\
\hline
Cora    & 2708 & 5429  & 0.0015  & 1433 \\
Citeseer    & 3312 & 4715 & 0.0009  & 3703 \\
PIT    & 851 & 16392 & 0.0227 & 1224 \\
\hline
\end{tabular}
\label{tab:data}
\end{table}

In the single-target setting, we follow Nettack~\cite{zugner2018adversarial} to consider the classification margin, i.e., the probability of ground-truth label minus the second-highest probability via the pre-train 2-layers GCN, and to select three different sets of target nodes for the experiment:
(a) high-confident targets: $10$ nodes with the highest positive margin of classification, (b) low-confident targets: $10$ nodes with the lowest positive margin of classification, and (c) Random targets: 20 more randomly-chosen nodes. 
The classification margin can exhibit both accuracy and confidence. The value of margin indicates the confidence level, and the positive/negative means correctly/wrongly classification.
We also follow Nettack to repeat the random splitting of train/validation/test up to five times, and report the average results. Besides, we set the maximum budget $b=20$.

For the multiple-target setting, all test nodes are considered as the targets.
The maximum budget is set as $b = 10$ for each target.
We repeat the experiment up tp $20$ times in terms of different data splitting, and report the average results for both perturbed nodes (denoted as ``Set'') and overall testing nodes (denoted as ``Overall''). 

Since no existing methods are on perturbing graphs for worsening PLC and maintaining TLC,
we consider the following baselines: 
(1) \textbf{clean}: generating the evaluation scores using the original graphs, i.e., no perturbation is performed;
(2) \textbf{pseudo-random perturbation} (RD): randomly perturbing candidate edges that pass the hypothesis test of unnoticeable power-law degree distribution (proposed by Nettack~\cite{zugner2018adversarial}); and 
(3) \textbf{adversarial attack} (NT): fully adopting Nettack~\cite{zugner2018adversarial},
which is similar to NetFense but considers only the PLC task, i.e., we apply Nettack's structured graph perturbations to lower down the PLC performance. 

\subsection{Single-Target Perturbations}
We display the results in Table~\ref{tab:single} and Fig.~\ref{fig:R1ST_cora} and~\ref{fig:R1ST_terr}. In each figure, the results of TLC and PLC are shown in the left and right, respectively.
The proposed NetFense method leads to more satisfying performance, comparing to RD and NT. NetFense maintains the classification margin in the TLC task, and lowers down the classification margin to approach zero in the PLC task. 
Specifically, in TLC, NetFense can keep the positive confidence in predicting the target label (i.e., close to clean), indicating that the perturbed graphs do not hurt the data utility. In PLC, NetFense can make both private labels (binary classification) tend to have equal prediction probabilities, which indicates the privacy can be protected.
NT largely reduces the classification margin of PLC. It is because NT directly perturbs the graphs to mislead the PLC task by remarkably imposing the opposite effect on the original data distribution. However,
the negative PLC margin generated by NT would result in the crisis of anti-inference since the task is binary classification.
In addition, RD gives both margins more moderate changes. Although the margins of TLC for RD keeps a similar level as clean in Cora and PIT, its PLC margin still holds at a high level, which exhibits the instability of RD.

We also present the averaged scores of margin and accuracy (Acc.), along with each method's performance difference from Clean, in Table~\ref{tab:single}.

\begin{figure}[!t]
  \centering
  \includegraphics[width=1.0\linewidth]{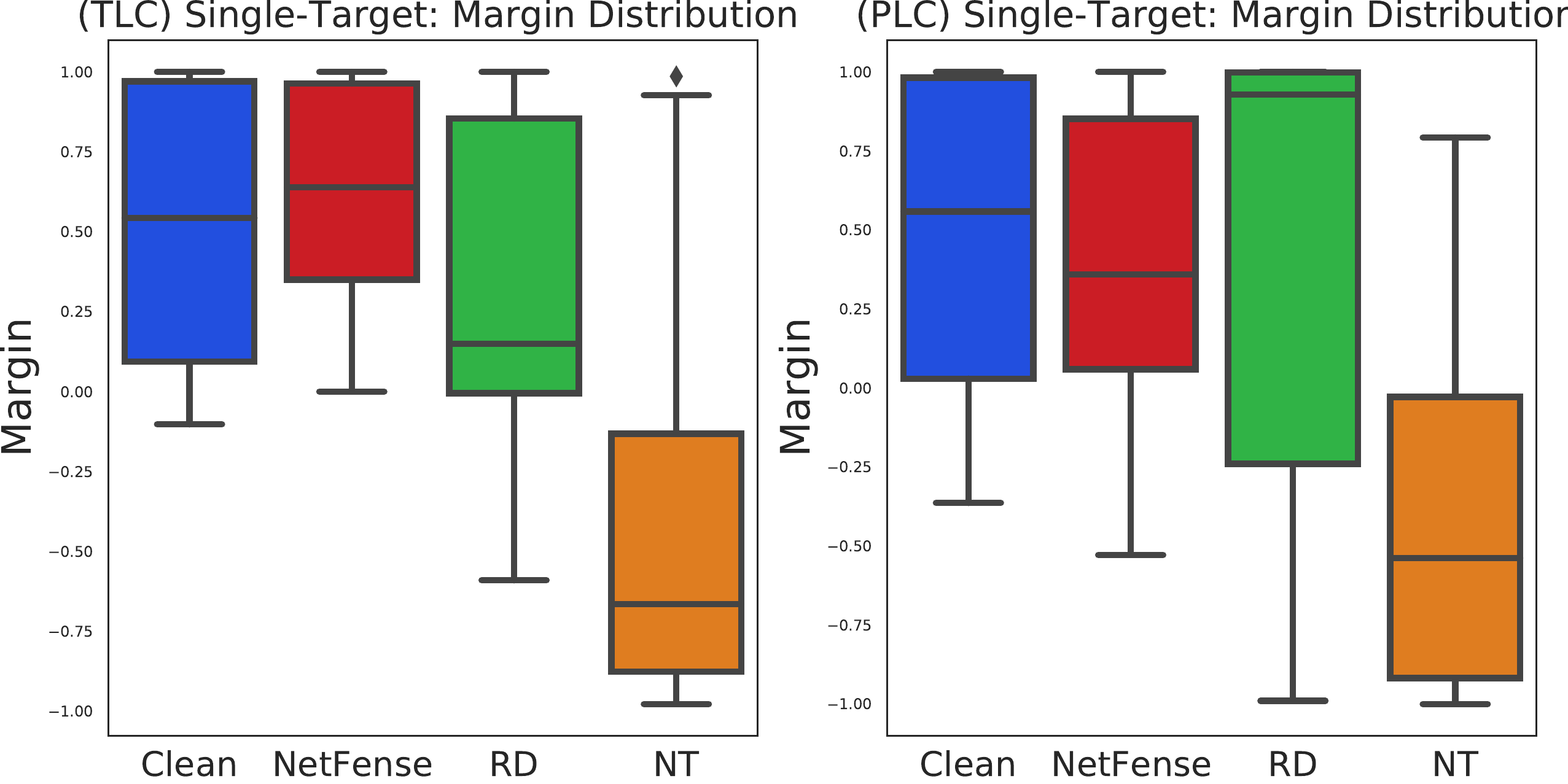}
  \caption{Boxplots of single targets' margin for Citeseer.}
  \label{fig:ST_cite}
\end{figure}

\begin{figure}[!t]
  \centering
  \includegraphics[width=1.0\linewidth]{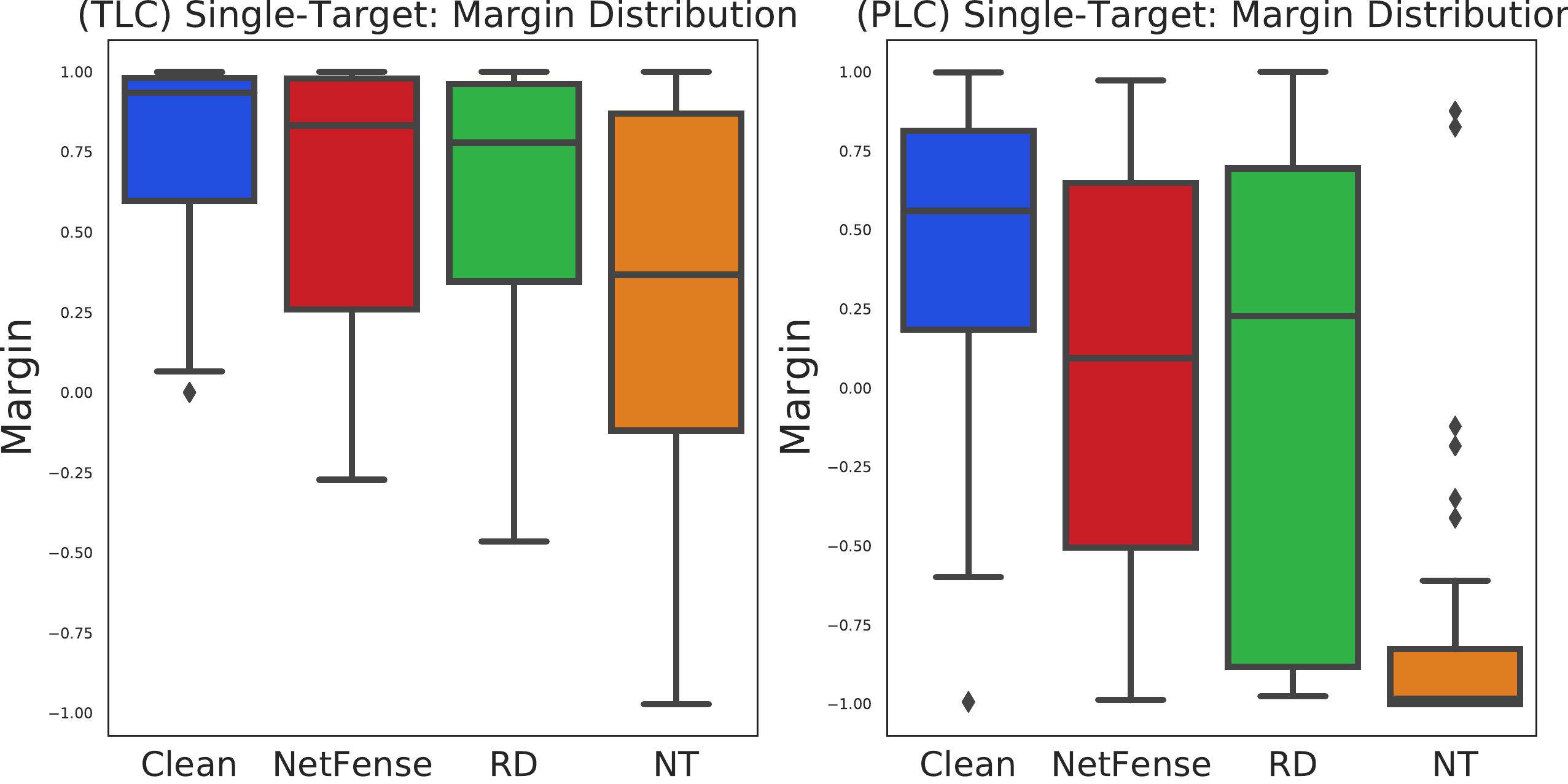}
  \caption{Boxplots of targets' margin for Cora.}
  \label{fig:ST_cora}
\end{figure}

\begin{figure}[!t]
  \centering
  \includegraphics[width=1.0\linewidth]{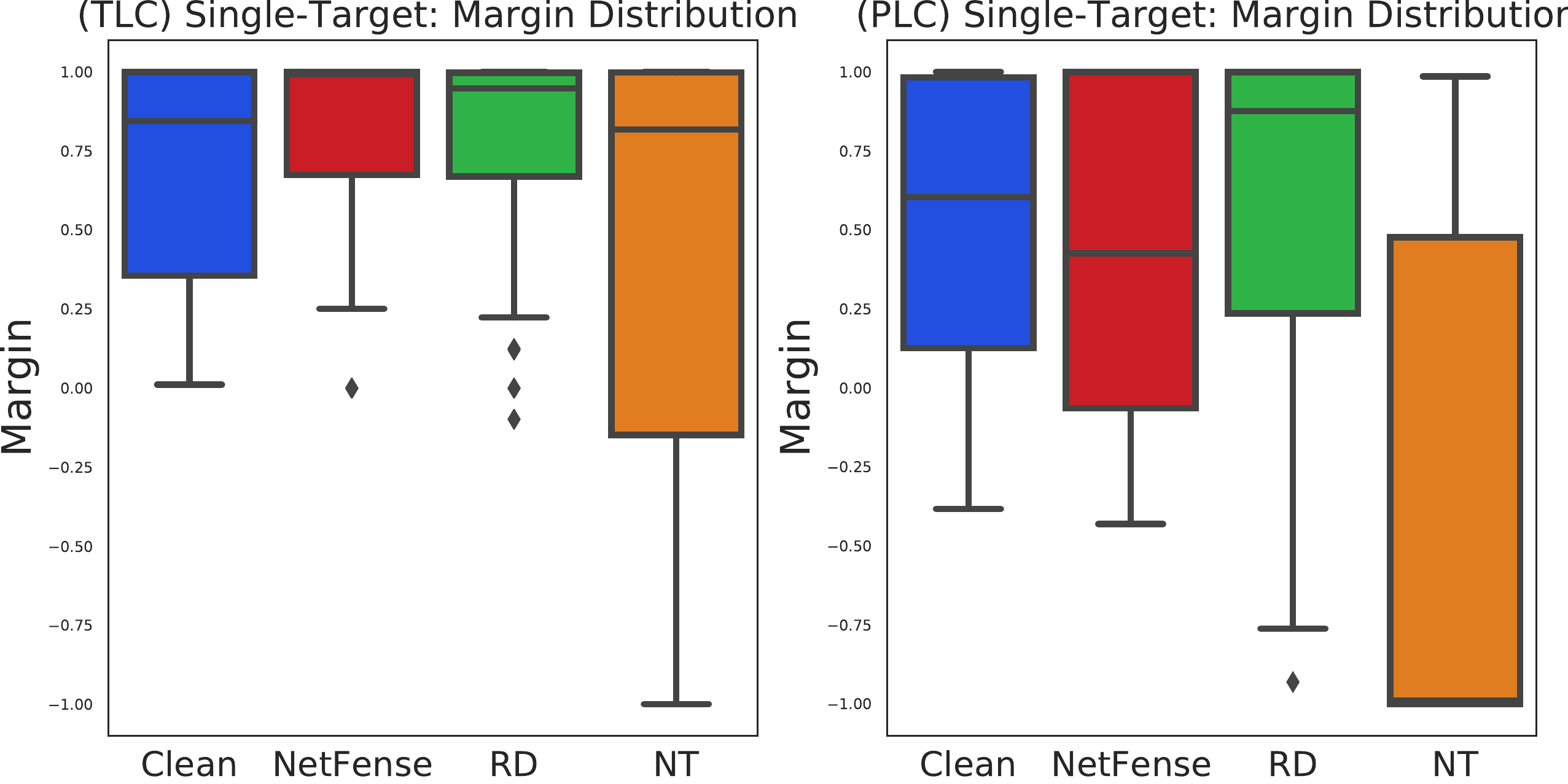}
  \caption{Boxplots of targets' margin for PIT.}
  \label{fig:ST_terr}
\end{figure}

Similar to the performance in Fig.~\ref{fig:ST_cite},~\ref{fig:ST_cora} and~\ref{fig:ST_terr}, 
the perturbation of RD and NT can neither well maintain TLC accuracy nor reduce PLC accuracy to approach $0.5$, comparing to our NetFense. In other words, NetFense can have TLC with smaller performance change, which preserves the data utility, and simultaneously make the accuracy of PLC binary classification much close to random prediction, which protects private labels.

\begin{table*}[!t]
\centering
\caption{Results of single-target perturbations.}
\resizebox{\textwidth}{!}{%
\begin{tabular}{l||c|rr|rr|rr|rr|rr|rr}
\hline
Methods                & Datasets   & \multicolumn{4}{c|}{Citeseer}      & \multicolumn{4}{c|}{Cora}          & \multicolumn{4}{c}{PIT} \\ \hline 
 &      & \multicolumn{2}{c|}{TLC (Margin/Acc.)}   & \multicolumn{2}{c|}{PLC (Margin/Acc.)}   & \multicolumn{2}{c|}{TLC (Margin/Acc.)}   & \multicolumn{2}{c|}{PLC (Margin/Acc.)}  & \multicolumn{2}{c|}{TLC (Margin/Acc.)}      & \multicolumn{2}{c|}{PLC (Margin/Acc.)} \\ \hline \hline
Clean & Score     & 0.545  & 0.980   & 0.616  & 0.840   & 0.735  & 0.960   & 0.278  & 0.685  & 0.657  & 1.000      & 0.544  & 0.825 \\ \hline
RD                     & Score     & {0.371} & {0.810} & 0.561          & 0.770          & 0.545         & 0.855         & 0.130            & 0.605         & {0.629} & 0.950          & 0.573          & 0.850           \\ 
                       & Difference & -0.174         & -0.170         & -0.055         & -0.070         & -0.190         & -0.105        & -0.148          & -0.080         & -0.028         & -0.050         & 0.029          & 0.025          \\ \hline
NT                     & Score     & -0.456         & 0.245         & -0.390          & {0.270} & 0.327         & 0.690          & -0.825          & {0.020} & 0.344          & 0.745         & -0.669         & {0.155} \\
                       & Difference & -1.001         & -0.735        & -1.006         & -0.570         & -0.408        & -0.270         & -1.103          & -0.665        & -0.313         & -0.255        & -1.213         & -0.670          \\ \hline
NetFense                   & Score     & \textbf{0.408}          & \textbf{0.845}         & \textbf{0.274} & \textbf{0.685}         & \textbf{0.596} & \textbf{0.915} & \textbf{0.045} & \textbf{0.540}          & \textbf{0.660}           & \textbf{0.955} & \textbf{0.284} & \textbf{0.635}          \\
                       & Difference & -0.137	& -0.135	& -0.342	& -0.155	& -0.139	& -0.045	& -0.233	& -0.145	& 0.003	& -0.045	& -0.260	& -0.190
     \\ \hline

\end{tabular}
}
\label{tab:single}
\end{table*}

\subsection{Multiple-Target Perturbations}
\begin{figure}[!t]
  \centering
  \includegraphics[width=1.0\linewidth]{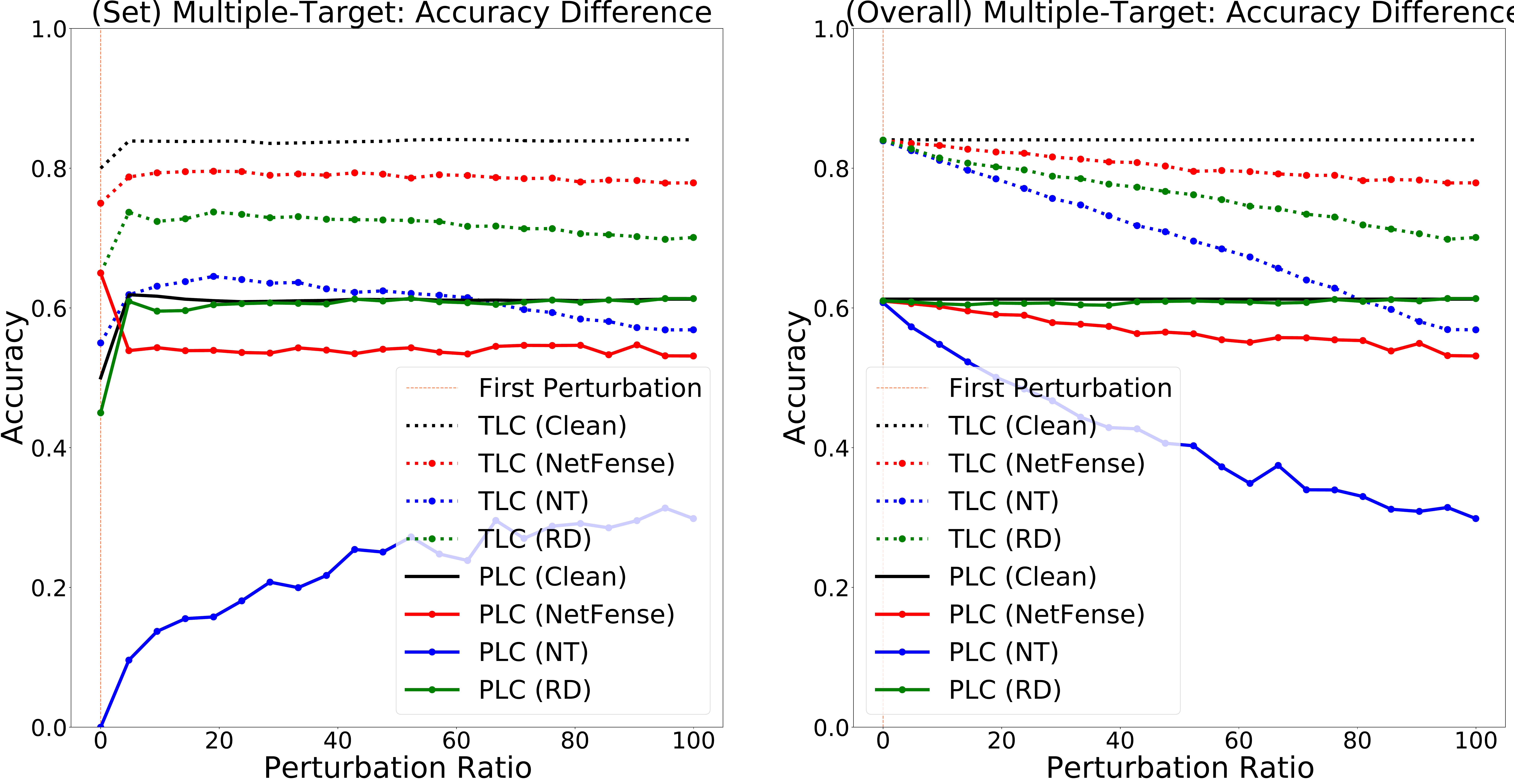}
  \caption{Multiple targets' accuracy for Cora.}
  \label{fig:MT_cora}
\end{figure}

\begin{figure}[!t]
  \centering
  \includegraphics[width=1.0\linewidth]{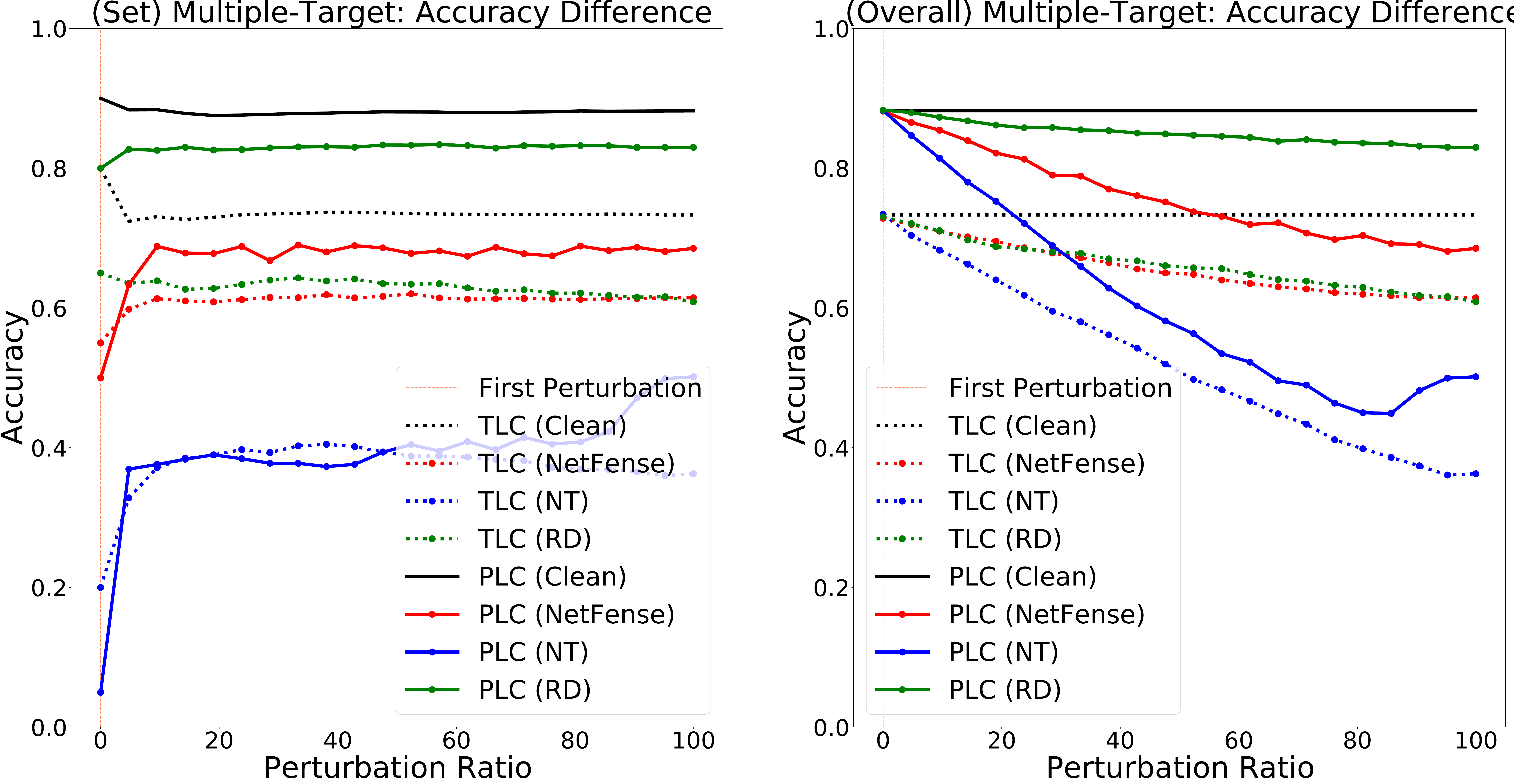}
  \caption{Multiple targets' accuracy for Citeseer.}
  \label{fig:MT_cite}
\end{figure}

\begin{figure}[!t]
  \centering
  \includegraphics[width=1.0\linewidth]{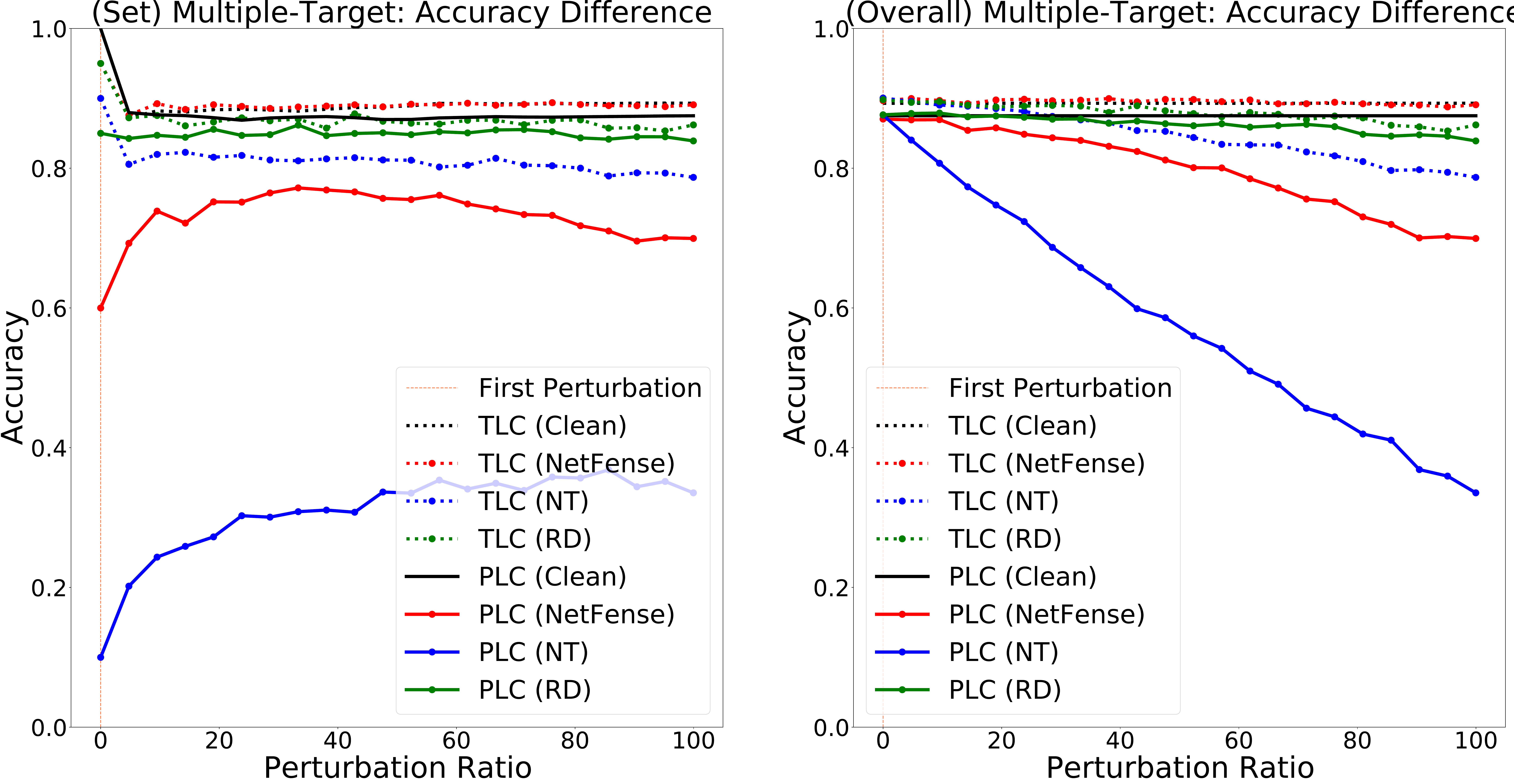}
  \caption{Multiple targets' accuracy for PIT.}
  \label{fig:MT_terr}
\end{figure}

We repeatedly apply the single-target perturbation on each testing node to have the results of multi-target perturbation.
The performance in terms of classification accuracy is exhibited at each time of single-target perturbation. By considering the perturbation ratio, i.e., the percentage of testing nodes being perturbed, as the x-axis, 
we show the results in Fig. \ref{fig:MT_cora}, \ref{fig:MT_cite} and \ref{fig:MT_terr}. Each figure exhibits the averaged accuracy of only perturbed nodes (``Set'') in the left and the accuracy of all testing nodes (``Overall'') in the right. 

Overall speaking, the proposed NetFense can better approach the TLC accuracy of Clean, along with stabler performance as the perturbation ratio increases, in both Set and Overall results. NetFense also simultaneously produces apparent accuracy drop in the task of PLC. Comparing to NT, although NT has stronger perturbation power to lower down the accuracy of PLC, we concern that the over-perturbation may lead to the crisis of anti-inference for binary private labels. That said, one could reverse the predicted labels to obtain the true private information. On the other hand, since NT does not consider to preserve the structural features for TLC, its perturbations apparently destroy the TLC accuracy. 

For ``Set'' accuracy, there exists obvious fluctuation when the perturbation ratio is low since fewer perturbed targets tend to result in extreme accuracy scores. After about $5\%~10\%$ perturbation ratio, the accuracy becomes stable. 
In addition, since NT imposes too much disturbance into the near-by graph structure of perturbed targets, 
its performance curves are more unstable than NetFense and RD. 
On the other hand, the curves of the ``Overall'' accuracy is decreasing as the increment of perturbation ratio. Such results exhibit that each target is gradually perturbed, and the perturbing effect is kept and even reinforced. 

\subsection{Advanced Experimental Analyses}
\label{sec-exp-adv}
\begin{figure}[!t]
  \centering
  \includegraphics[width=1.0\linewidth]{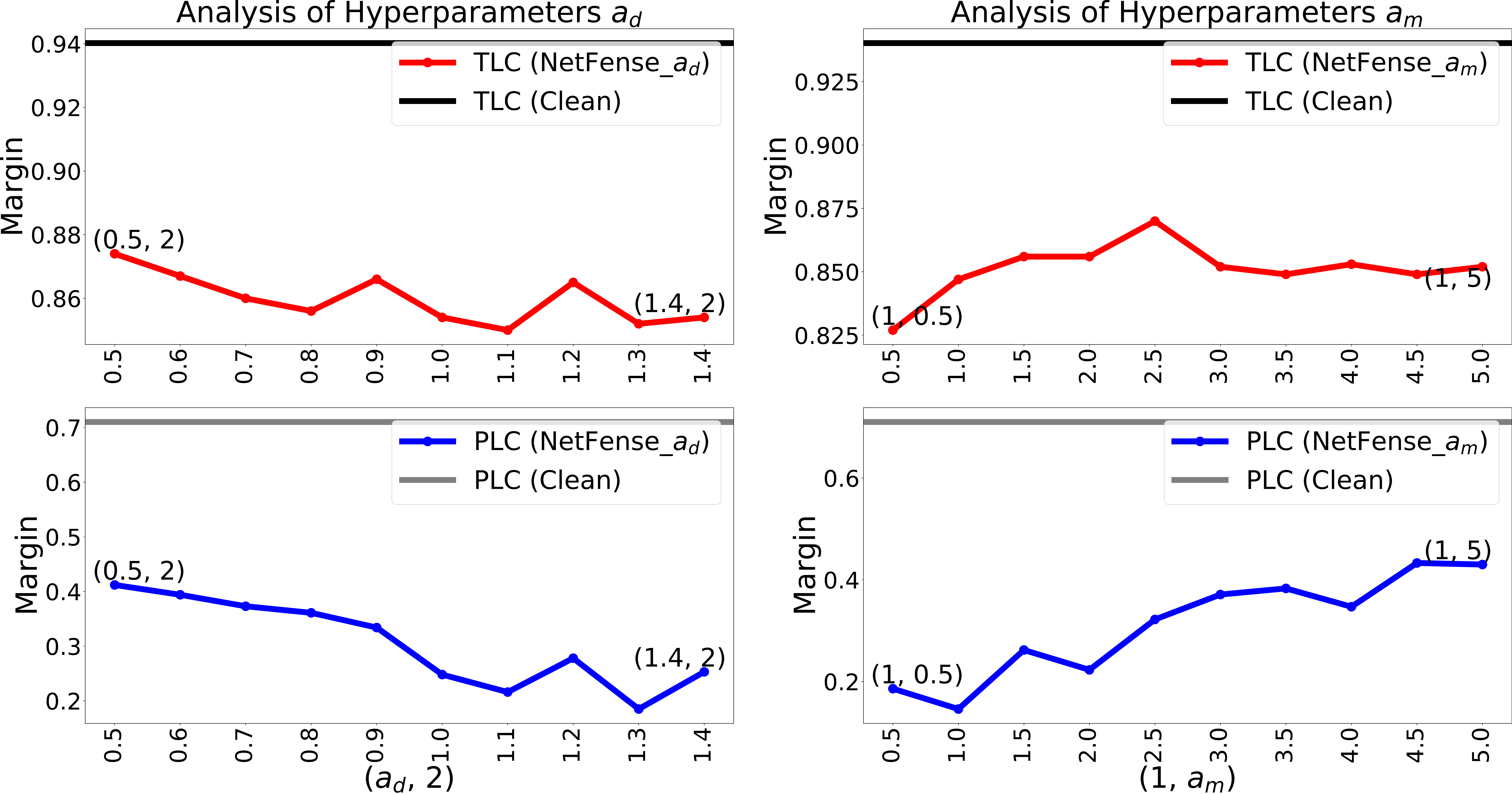}
  \caption{Effects of ${a_d}$ and ${a_m}$ in PIT dataset.}
  \label{fig:Hyper}
\end{figure}

\textbf{Balancing Factors of TLC and PLC.} 
By performing single-target perturbation using our NetFense, we examine the effect of the hyperparameters ${a_d}$ and ${a_m}$ in Eq.~\ref{eq:Score} that determines the balance between TLC and PLC. As the results on the other datasets exhibit similar trends, we only show the results on the PIT dataset due to the page limit.
Without loss of generality, we discuss the effect of positive values of ${a_d}$ and ${a_m}$. 
We fix one and present the effect of the other by varying ${a_d}$ and ${a_m}$ around $1$. The results in terms of TLC and PLC classification margin are exhibited in Fig.~\ref{fig:Hyper}.
In the left, we can find both TLC and PLC performance is quite sensitive to ${a_d}$. It is because ${a_d}$ is directly related to the perturbation goal. Higher $a_d$ drops PLC performance more significant than TLC performance. We choose higher $a_d$ as it makes NetFense to select perturbation candidates that heavily hurt PLC but slightly hurt TLC.
In the right of Fig.~\ref{fig:Hyper}, the results of varying $a_m$ are opposite to those of $a_d$. Increasing $a_m$ strengthens the maintenance of TLC performance but weakens the privacy protection effect of PLC (i.e., raising the margin towards that of Clean). 
Higher $a_m$ leads NetFense to avoid selecting candidates that can hurt TLC.
Nevertheless, since there could exist some features shared by both TLC and PLC, too stronger maintenance power (i.e., higher $a_m$) can limit the perturbation capability, i.e., increasing the margin of PLC. Eventually we suggest $(a_d, a_m)=(2, 1)$ can better maintain TLC performance and decrease PLC margin.
\comments{
Therefore, we keep $a_d = 1$ as the original value and then select $a_m$ to ensure an appropriate effect of maintenance. We choose $a_m = 2$ because the red curve of TLC start to convergence around $a_m = 2$ (top - right) and its corresponding margin of PLC is lower. Additionally, we find the margins of $a_d = 2$ with fixed $a_m$ for PLC and TLC can still hold a medium level.
}

\begin{figure}[!t]
  \centering
  \includegraphics[width=1.0\linewidth]{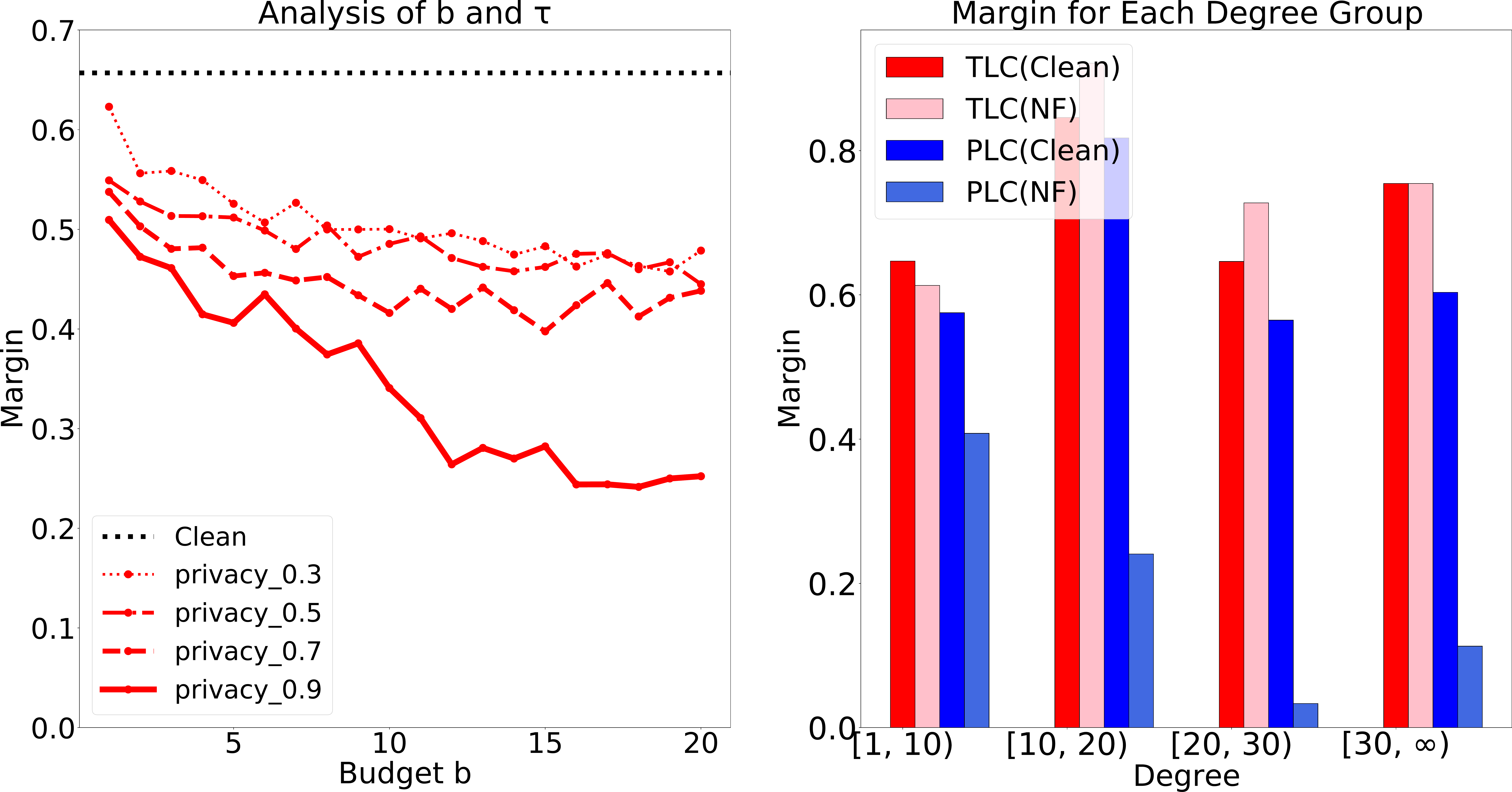}
  \caption{Effects of perturbation factors (i.e., $b$ and $\tau$) on PLC (left) and node degree on TLC and PLC (right) in PIT.}
  \label{fig:D_N}
\end{figure}

\textbf{Perturbation Factors.} 
We demonstrate how the perturbation factors, including the perturbation budget $b$, the threshold $\tau$ of data unnoticeability, and node degree, affect the performance.
We conduct the single-target perturbation with various $(b,\tau)$ combinations
on the PIT dataset, and show the PLC margin that reflects privacy disclosing. 
We select $\tau$ via the quantile $q\in \{0.3, 0.5, 0.7, 0.9\}$ of $\Delta_{u \leftrightarrow  v}P_{g}$ for all $u, v \in V$, denoted as $privacy_q$.
The results shown in Fig.~\ref{fig:D_N} (left) exhibit that higher budget $b$ leads to better privacy protection (lower PLC margin towards zero). It is because a higher $b$ brings more perturbations.
The steep decrements of PLC margins indicate most of the edges that could reveal private labels are perturbed. Higher $\tau$ leads to better privacy protection (i.e, steeper decrements). Nevertheless, there is a trade-off between data unnoticeablility and privacy protection. Here we pay more attention to privacy protection, and thus apply higher $\tau$ (i.e., $q = 0.9$) to consider more candidates being perturbed.
A higher $\tau$ (i.e., higher $q$) can help identify the edges whose perturbations lead to a nearly unnoticeable graph change, and prevent significant edges from being the candidates.

\textbf{Effect of Node Degree.}
Since high-degree nodes have more neighbors to learn their embeddings, they might have higher potential being influenced by perturbations. Here we aim at investigating how node degree correlates with the performance of TLC and PLC after perturbations. We conduct single-target perturbations on PIT data, and compare the results between Clean and NetFense (NF). Nodes are divided into four groups based on their degree values. The results are exhibited in Fig.~\ref{fig:D_N} (right).
For TLC, we can find the margins do not change too much across after NF perturbations various degree ranges. That said, NetFense can maintain TLC performance for both high- and low-degree nodes.
For PLC, the decrements after NetFense perturbations are more obvious.
Moreover, the difference of margin between Clean and NetFense becomes significant as the degree increases
(i.e., $0.167, 0.577, 0.532, 0.491$).
Such results deliver that the privacy protection capability of NetFense is quite significant for high-degree nodes because the PLC margin can be reduced to approach zero.   

\textbf{Strategies of Selecting Data-Unnoticeable Candidates.} 
NetFense relies on the revised PPR for data-unnoticeable candidate edge selection in Sec.~\ref{sec-cand}. Here we aim to examine whether the perturbed graphs according to our proposed revised PPR can truly keep the local neighborhood of each node unchanged. We think a good strategy of selecting data-unnoticeable candidates should hurt the local structure of every node as minimum as possible in the perturbed graphs. We compare the proposed revised PPR-based selection (i.e., Eq.~\ref{eq:oneside}) with three baseline strategies of candidate selection. 
(a) \textit{Random Choice}: randomly selecting edges without replacement.
(b) The hypothesis testing of degree distribution~\cite{clauset2009power}, which is adopted by Nettack~\cite{zugner2018adversarial}: we examine the scaling parameter of power-law distribution
based on the likelihood ratio test for the hypotheses $\{$ $H_0$: distributions of $G$ and $G'$ are the same$\}$ and $\{$ $H_1$: distributions of $G$ and $G'$ are different from each other$\}$, where the graph $G'$ indicates the graph with at least one edge perturbation from $G$. We identify the unnoticeability of an edge perturbation if the likelihood ratio is lower than a threshold $\chi_{(1),0.95}^2 \approx 0.004$ for the given significance level to p-value $=0.05$. We determine the edge with the smallest ratio as the candidate.
(c) \textit{PPR (Original)}: according to our discussion (i.e., Eq.~\ref{eq:oneside} and Eq.~\ref{eq:twoside})
in Sec.~\ref{sec-cand}, we can derive the influence of edge perturbation via the original PPR score for the version without the direction of adjustment, given by: $\Delta_{u \to v} P^{o}_{g}((i \to j))= \sum_{i} c' (M_1^{-1})_{iu} /(1-c'(M_1^{-1})_{vu})$). The edges with lower scores are considered as good unnoticeable candidates. 
\comments{
Note that the difference between PPR (Original) and our revised PPR lies in
their denominators of (Eq.\ref{eq:oneside} vs. Eq.~\ref{PPR-revised}), which reverses the sign and the direction of the PPR score of $v,u$. The trick helps us to make the low influence edge with a higher score to remove, and we increase the score to add the edge with high PPR.
}
To examine which strategy can better maintain the local graph neighborhood, we utilize \textit{average of local clustering coefficient}, denoted by \textbf{CA}, to be the evaluation metric. For each strategy, we sequentially perturb the candidates in the descent order of their unnoticeable scores, and report the CA score at a time.
The damage degree of local neighborhoods resulting from the accumulated edge perturbations can be reflected by the decrement of CA. 

\begin{figure}[!t]
  \centering
  \includegraphics[width=1.0\linewidth]{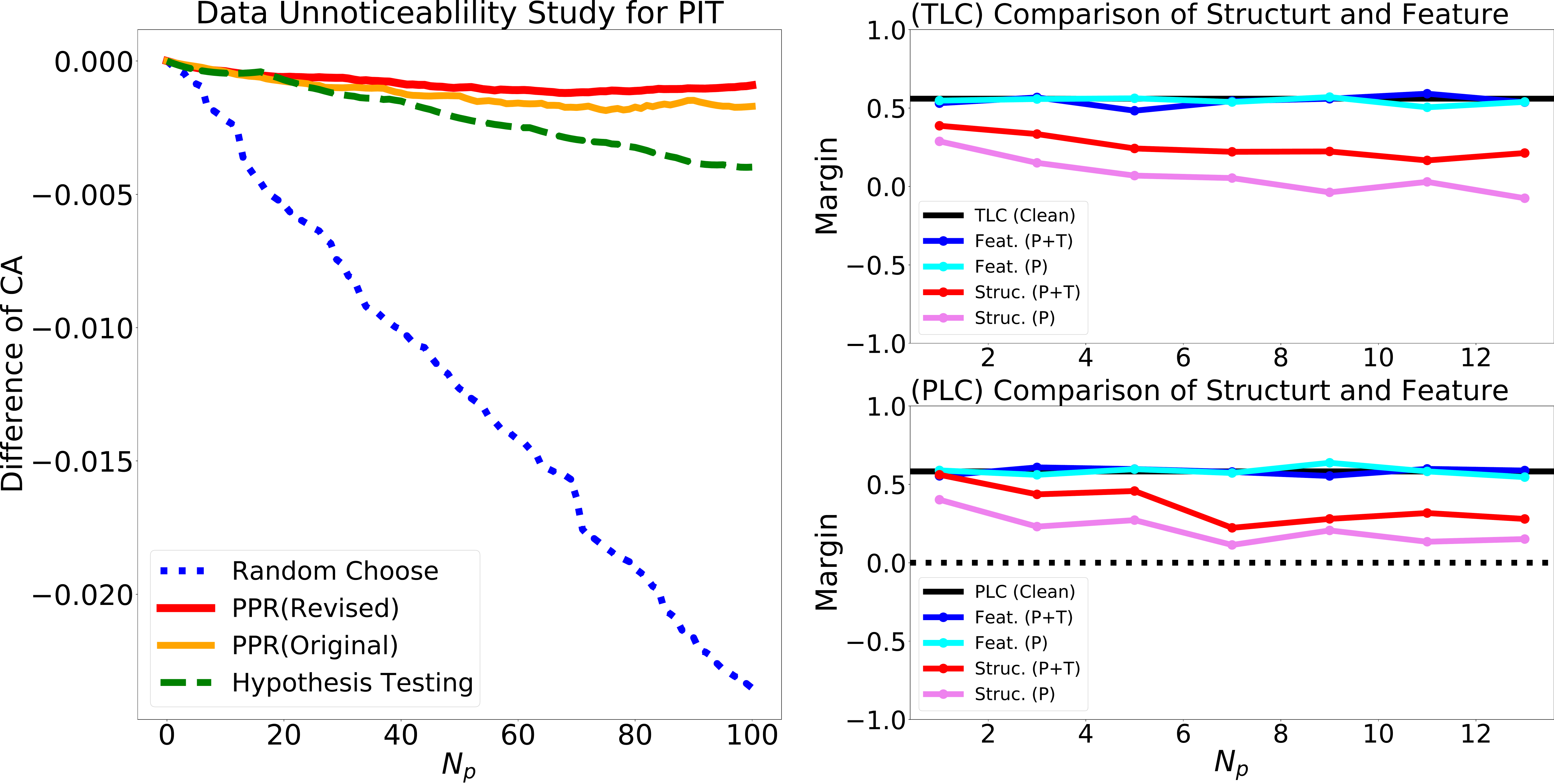}
  \caption{Left: effect of different data-unnoticeable candidate selection strategies in PIT. Right: comparison of structure and feature effects in Citeseer.}
  \label{fig:Un_FS}
  \vspace{-1em}
\end{figure}

We present the results on the PIT dataset in Fig.~\ref{fig:Un_FS} (left), in which x-axis is the number of perturbed edges ($N_p$), and y-axis is the difference between CA scores before and after the perturbations. 
We can apparently find that PPR-based strategies demonstrate promising preservation of local graph structure (i.e., lower CA differences). The proposed revised PPR can have almost unchanged CA scores. 
Random Choice tremendously reduces CA scores. The hypothesis testing of degree distribution alleviates the destruction of local structure, but still cannot well maintain local connectivity because it considers no interactions between nodes that are exploited by PPR-based strategies.

\textbf{Structure vs. Feature Perturbations.}
Recall that in Sec.~\ref{sec-infgnn} we have discussed that perturbing graph structure (\textbf{Struc.}) brings a more significant influence on GNN than perturbing node features (\textbf{Feat.}). 
Now we aim to empirically examine and compare structural and feature perturbations for TLC and PLC margins under single-target perturbation.
To have a fair comparison, either all edges or all nodes' features are considered for perturbations by NetFense. That said, we do not perturb candidate selection here. 
Since perturbing node features can affect the feature matrix in GNNs, we keep $\hat{A}$ but replace $X$ by the perturbed feature matrix $X'$ for the propagation term (i.e., $\hat{A}^2XW$) in the loss function Eq.~\ref{eq:Score}.
To further understand whether perturbing node features are effective in protecting private labels, the analysis is conducted via two cases: (1) maintaining TLC performance and reducing PLC prediction confidence by setting $a_d>0$ and $a_m>0$, denoted by \textbf{P+T}, and (2) only reducing PLC prediction confidence by setting $a_d>0$ and $a_m=0$, denoted by \textbf{P}, which implies protecting private labels without preserving data utility. 

By reporting the margin from accumulated perturbations of edges or node features, we show the results in Fig.~\ref{fig:Un_FS} (right) in Citeseer. 
We can find that the feature perturbations in both settings of P+T and P apparently maintain the TLC margins as Clean, but fails to reduce the PLC margins. That said, perturbing node features cannot effectively prevent private labels from being inferred.
The structural perturbations sacrifice some TLC margins (i.e., decreases TLC margins) for lowering down the PLC margins so that the private labels can be protected. 
In addition, to maintain data utility, the P+T case of structural perturbation cannot work that well as the P case. 
These results indicate that perturbing node features can only maintain data utility in terms of TLC, and has no effect in privacy protection.
Perturbing edges with having TLC in the loss function (i.e., P+T) is able to strike the balance between keeping data utility and preventing private labels from being confidentially inferred.

\begin{figure}[!t]
  \centering
  \includegraphics[width=1.0\linewidth]{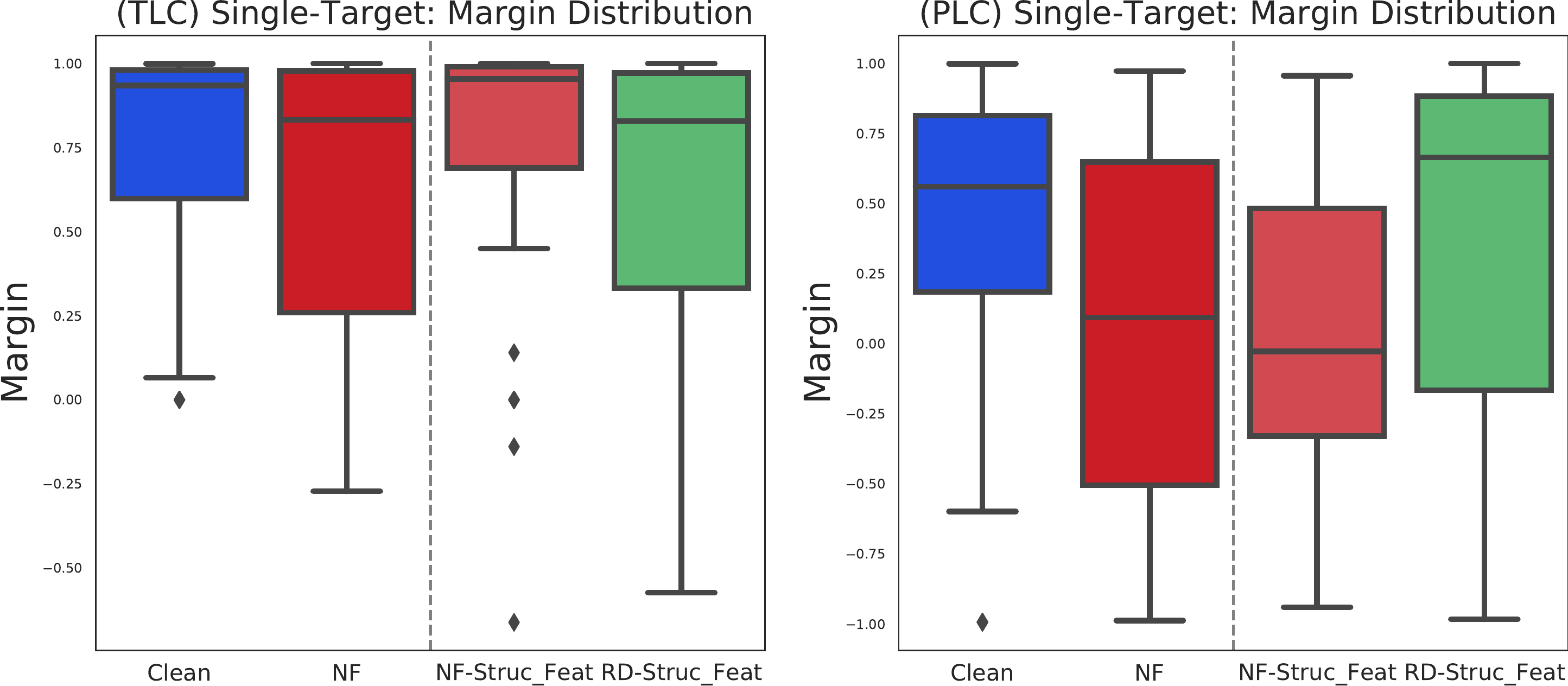}
  \caption{(Struct+Feat) Boxplots of targets' margin for Cora.}
  \label{fig:SFST_cora}
\end{figure}

\begin{figure}[!t]
  \centering
  \includegraphics[width=1.0\linewidth]{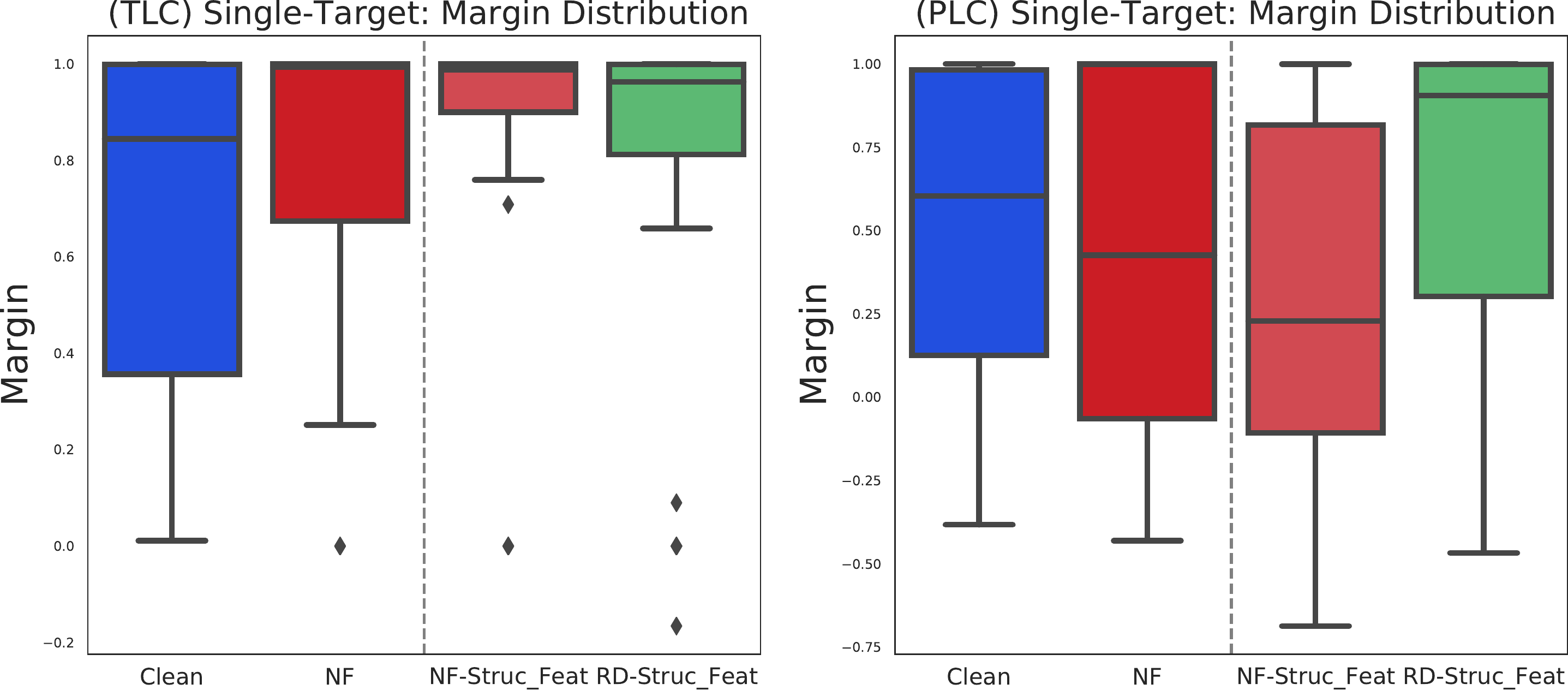}
  \caption{(Struct+Feat) Boxplots of targets' margin for PIT.}
  \label{fig:SFST_terr}
\end{figure}

\ct{
\textbf{Perturbing Both.}
We discuss whether perturbing both feature and structure can bring more effective results of NetFense. Single-target perturbation with NetFense (NF) and Random (RD) are considered.
We follow Nettack~\cite{zugner2018adversarial} to combine both perturbation options by choosing either structure or feature with the higher score for each time of perturbation. The score is defined by the gradient-based function in Eq.~\ref{eq:Score}. That said, there could be some structural and some feature perturbations at the end of perturbation process.
Note that RD is to randomly choose either structure or feature for each time of perturbation with uniform probabilities. 
The results are exhibited in Fig.~\ref{fig:SFST_cora} and~\ref{fig:SFST_terr} for Cora and PIT datasets, respectively, in which NF-Struct$\_$Feat and RD-Struct$\_$Feat denote the results of NetFense and Random, respectively.
We can find that NetFense with perturbing both structure and feature leads to higher-margin values in TLC and approaches to zero confidence in PLC, and the results look a bit better than NetFense with perturbing only structure.
Random (RD-Struct$\_$Feat) perturbation does not consider how to maintain TLC performance and decrease PLC confidence, and thus produces worse results. 
Although perturbing both is more effective for TLC and PLC, we still need to point out that perturbing features has two concerns. First, as discussed in Sec.~\ref{sec-infgnn}, perturbing features bring much more computation cost. Second, it is less reasonable to perturbing features (i.e., modifying user profiles) by an algorithm in online social platforms because they should be manually filled and modified by users. Therefore, given the performance of perturbing both and perturbing only structure is competitive, we suggest perturbing only structure for real applications.
}


\begin{figure}[!t]
  \centering
  \includegraphics[width=1.0\linewidth]{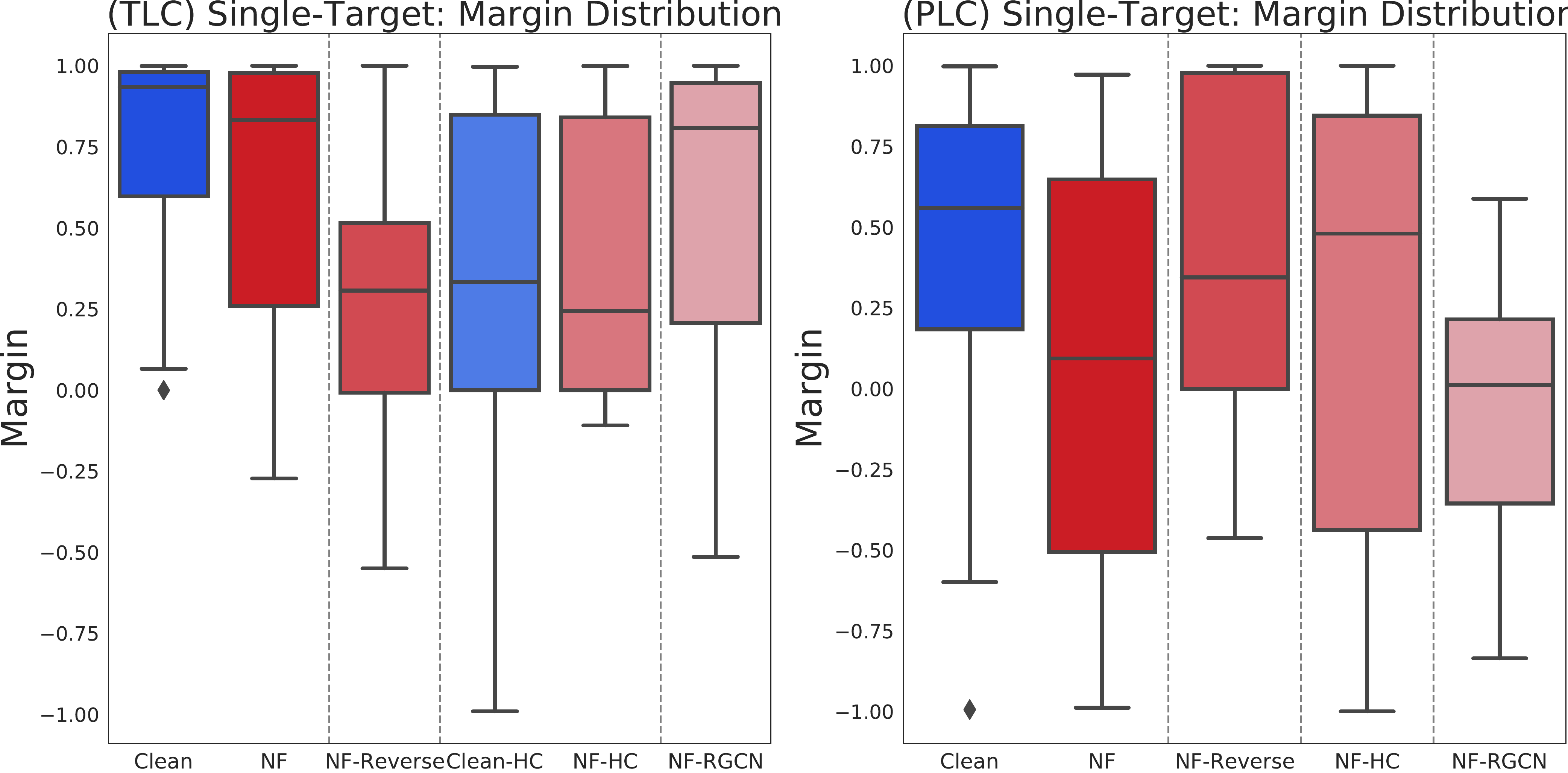}
  \caption{Boxplots for Cora on various settings.}
  \label{fig:R1ST_cora}
\end{figure}

\begin{figure}[!t]
  \centering
  \includegraphics[width=1.0\linewidth]{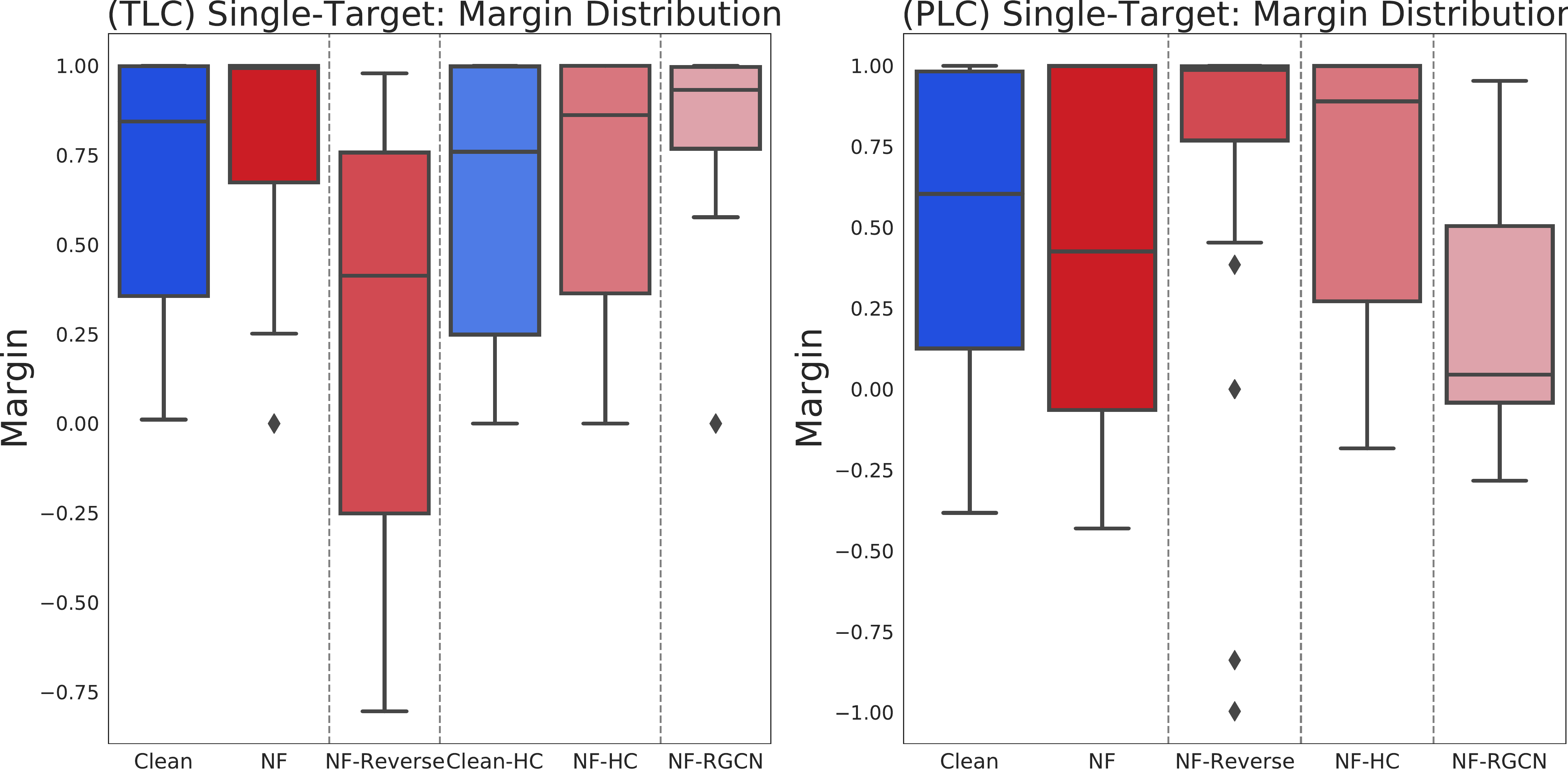}
  \caption{Boxplots for PIT on various settings.}
  \label{fig:R1ST_terr}
\end{figure}

\ct{
\textbf{Label Swapping.}
To understand whether the proposed NetFense can be also effective for different classification tasks, we swap the classification tasks of PLC and TLC, i.e., considering the original private labels as targeted labels and the original targeted labels as private labels, and conduct the same experiments. The results are displayed in Fig.~\ref{fig:R1ST_cora} and~\ref{fig:R1ST_terr} (NF-Reverse) for Cora and PIT datasets, respectively, in which the ``Reverse'' results indicate the performance when PLC and TLC labels are swapped. We can find that our NetFense can successfully lower down the prediction confidence of TLC (as it is now considered to be protected) and simultaneously keep the high performance of PLC (as it is now considered to be maintained). Such results also verify that NetFense can protect any specified labels from being confidentially attacked and maintain the performance of any specified targeted labels.}

\ct{
\textbf{Label Correlation.}
We aim at studying how the correlation between PLC and TLC labels affects the performance of our NetFense. We create a new TLC label that is highly correlated with the PLC label. While the label of PLC is binary, suppose the number of labels for TLC is $\vartheta$, the new TLC label of node $v_i$ is defined as: its new TLC label $c'_i$ is the same as the original TLC label $c_i$, i.e., $c'_i = c_i$, if its PLC label is $p_i = 1$; otherwise, if its PLC label is $p_i = 0$, we set $c'_i$ to be an extra newly-created label (i.e., the ($\vartheta+1$)-th label).
With Cohen's kappa coefficient, we display the correlation change between TLC and PLC labels before and after modifying the TLC labels.
\comments{To examine the correlation change between TLC and PLC labels before and after modifying the TLC label, 
we adopt \textit{Cohen's kappa coefficient} $\kappa$ value \footnote{\url{https://en.wikipedia.org/wiki/Cohen\%27s_kappa}}, which can measure the correlation of two discrete distributions. Higher $\kappa$ (close to 1) indicates more similar agreements, and lower $\kappa$ (close to -1) means less consistent than random selection.} 
We find that $\kappa$ coefficients between TLC and PLC labels change from $0.01$ and $-0.14$, to $0.33$ and $0.63$ for 
Cora, and PIT, respectively. Such $\kappa$ differences imply that the new TLC label is highly correlated to the PLC label in this experiment. 
We present performance of Clean and NetFense for the new TLC with the new ``Higher Correlated'' (HC) label
in Fig.~\ref{fig:R1ST_cora} and~\ref{fig:R1ST_terr}. The new TLC performances without perturbation (Clean-HC) are similar to the scores of the original TLC. The new TLC scores on NetFense (NF-HC) can be maintained as the original level; however, the PLC scores are a bit worse than the original setting (training with the original TLC label). Such results tell us that when the PLC label is highly correlated with the TLC label, it would become more challenging for NetFense to maintain the effectiveness of PLC. Since accurately classifying the mutually-correlated TLC and PLC labels involves learning similar information from the data, the model naturally cannot better distinguish them from each other during the process of adversarial-defense perturbation.
}


%



\ct{
\textbf{Against Robust GNN.}
We further aim to study whether the perturbed/defended graphs can survive from robust GNNs on PLC and TLC tasks because the adversary can choose to utilize robust GNN as a stronger attacking model to avoid perturbation influence and disclose user privacy. In other words, we examine how robust are the perturbed graphs generated by NetFense against robust GNNs. One may expect that robust GNNs can generate better prediction performance on TLC and PLC. We employ RobustGCN (RGCN)~\cite{zhu2019robust} as the robust attacker, which has no knowledge about whether the graph is perturbed. A two-layer RCGN is implemented and performed on the perturbed graphs generated by single-target NetFense, denoted as NF-RGCN. The results are reported in Fig.~\ref{fig:R1ST_cora} and~\ref{fig:R1ST_terr}. We can find that on the TLC task, the performance of RGCN (i.e., NF-RGCN) is competitive with that of our simplified GCN (i.e., NF). In more details, NF is slightly better to maintain the TLC performance. On the other hand, for the PLC task, RGCN (i.e., NF-RGCN) leads to lower classification confidence than our simplified GCN (i.e., NF). Such results further prove the usefulness of our NetFense model because even the robust GNN (RGCN) cannot increase the prediction confidence on the PLC task, i.e., the private labels are still well protected by NetFense. That said, the perturbed graphs generated by NetFense are verified to have solid privacy protection against robust GNN attackers.
}

\section{Conclusions}
\label{sec-concl}
This paper presents a novel research task: adversarial defenses against privacy attack via graph neural networks under the setting of semi-supervised learning. 
We analyze and compare the differences between the proposed problem and model attacks on graph data, and realize the perturbed graphs should keep data unnoticeability, maintain model unnoticeability (i.e., data utility), and achieve privacy protection at the same time.
We develop an adversarial approach, NetFense, and empirically find that the graphs perturbed by NetFense can simultaneously lead to the least change of local graph structures, maintain the performance of targeted label classification, and lower down the prediction confidence of private label classification. We also exhibit that perturbing edges brings more damage in misclassifying private labels than perturbing node features. In addition, the promising performance of the proposed NetFense lies in not only single-target perturbations, but also multi-target perturbations that cannot be well done by model attack methods such as Nettack. The evaluation results also deliver that 
moderate edge disturbance can influence the graph structure to avoid the leakage of privacy via GNNs and alleviate the destruction of graph data. 
Besides, we also offer the analysis of hyperparameters and perturbation factors that are highly related to the performance. We believe the insights found in this study can encourage future work to further investigate how to devise a privacy-preserved graph neural networks, and to study the correlation between the leakage of multiple private labels and attributed graphs.
\vspace{-5pt}


\ifCLASSOPTIONcompsoc
  \section*{Acknowledgments}
\else
  \section*{Acknowledgment}
\fi

This work is supported by Ministry of Science and Technology (MOST) of Taiwan under grants 109-2636-E-006-017 (MOST Young Scholar Fellowship) and 109-2221-E-006-173, and also by Academia Sinica under grant AS-TP-107-M05.

\bibliographystyle{plain}
\bibliography{NetFense}

\vspace{-40pt}
\begin{IEEEbiography}[{\includegraphics[width=1in,height=1.25in,clip,keepaspectratio]{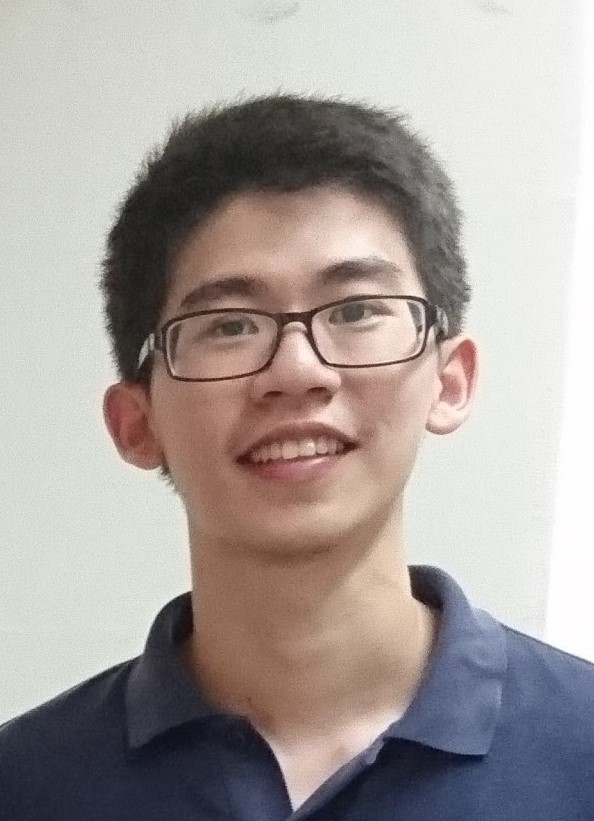}}]%
{I-Chung Hsieh} received a Master's degree in Science in Statistical Science from National Chung Cheng University, Chiayi, Taiwan, in 2017. He is a research assistant in Networked Artificial Intelligence Laboratory at National Cheng Kung University, Tainan, Taiwan from 2018. His main research interests include Graph Representation Learning, Graph Neural Networks, Data Science, Machine Learning, and Deep Learning. He had several papers published at top venues, including IEEE TKDE 2021 and NeurIPS GRL 2019.
\end{IEEEbiography}
\vspace{-35pt}
\begin{IEEEbiography}[{\includegraphics[width=1in,height=1.25in,clip,keepaspectratio]{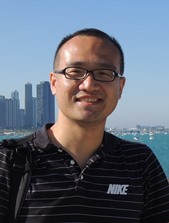}}]%
{Cheng-Te Li} is an Associate Professor at Institute of Data Science and Department of Statistics, National Cheng Kung University (NCKU), Tainan, Taiwan. He received my Ph.D. degree (2013) from 
National Taiwan University. 
Dr. Li's research targets at Machine Learning, Data Mining, Social Networks, Recommender Systems, and Natural Language Processing. He has papers published at top conferences, including KDD, WWW, ICDM, CIKM, SIGIR, IJCAI, ACL, EMNLP, NAACL, and ACM-MM. He leads Networked Artificial Intelligence Laboratory (NetAI Lab) at NCKU.
\end{IEEEbiography}
\end{document}